%% file: main.tex
\documentclass[10pt,twocolumn,letterpaper]{article}

\usepackage{wacv}              

\usepackage{graphicx}
\usepackage{amsmath}
\usepackage{amssymb}
\usepackage{booktabs}
\usepackage[table,xcdraw]{xcolor}
\usepackage{makecell}
\usepackage{multicol}
\usepackage{multirow}
\usepackage{siunitx}
\usepackage{balance}
\usepackage[table]{xcolor}
\usepackage[pagebackref,breaklinks,colorlinks]{hyperref}

\usepackage[capitalize]{cleveref}
\crefname{section}{Sec.}{Secs.}
\Crefname{section}{Section}{Sections}
\Crefname{table}{Table}{Tables}
\crefname{table}{Tab.}{Tabs.}

\def\wacvPaperID{1426} 
\def\confName{WACV}
\def\confYear{2025}

\begin{document}

\title{Forensic Iris Image-Based Post-Mortem Interval Estimation}

\author{Rasel Ahmed Bhuiyan\\
University of Notre Dame\\
Notre Dame, 46556, Indiana, USA\\
{\tt\small rbhuiyan@nd.edu}
\and
Adam Czajka\\
University of Notre Dame\\
Notre Dame, 46556, Indiana, USA\\
{\tt\small aczajka@nd.edu}
}
\maketitle

\begin{abstract}
   \input{sections/abstract}
\end{abstract}

\section{Introduction}
\input{sections/introduction}

\section{Related Works}
\input{sections/related-works}

\section{Methodology}
\input{sections/methodology}

\section{Experiments and Results}
\input{sections/experiment}

\section{Discussion}
\input{sections/discussion}

\balance
{\small
\bibliographystyle{ieee_fullname}
\bibliography{egbib}
}

\clearpage
\appendix
\renewcommand\thefigure{A\arabic{figure}}
\setcounter{figure}{0} 
\setcounter{section}{0}
\input{sections/supplementary}

\end{document}

%% file: sections/abstract.tex
Post-mortem iris recognition is an emerging application of iris-based human identification in a forensic setup. One factor that may be useful in conditioning iris recognition methods is the tissue decomposition level, which is correlated with the post-mortem interval (PMI), \ie the number of hours that have elapsed since death. PMI, however, is not always available, and its precise estimation remains one of the core challenges in forensic examination. This paper presents the first known to us method of the PMI estimation directly from iris images captured after death. To assess the feasibility of the iris-based PMI estimation, we designed models predicting the PMI from (a) near-infrared (NIR), (b) visible (RGB), and (c) multispectral (RGB+NIR) forensic iris images. Models were evaluated following a 10-fold cross-validation, in (S1) sample-disjoint, (S2) subject-disjoint, and (S3) cross-dataset scenarios. We explore two data balancing techniques for S3: resampling-based balancing (S3-real), and synthetic data-supplemented balancing (S3-synthetic). We found that using the multispectral data offers a spectacularly low mean absolute error (MAE) of $\approx 3.5$ hours in the scenario (S1), a bit worse MAE $\approx 17.5$ hours in the scenario (S2), and MAE $\approx 45.77$ hours in the scenario (S3). Additionally, supplementing the training set with synthetically-generated forensic iris images (S3-synthetic) significantly enhances the models' ability to generalize to new NIR, RGB and multispectral data collected in a different lab. This suggests that if the environmental conditions are favorable (\eg, bodies are kept in low temperatures), forensic iris images provide features that are indicative of the PMI and can be automatically estimated.

%% file: sections/introduction.tex

Iris recognition, after demonstrating its capabilities to serve as a biometric identifier up to a few weeks after death, and with development of government-led datasets of iris images, such as the FBI’s Next Generation Identification \cite{FBI_NGI_webpage},  
it has been observing an increased interest in forensic applications. Iris recognition can serve as an element of the forensic toolkit, aiding in cold case resolutions, or a standalone method allowing for quick and accurate registering bodies in mass fatality incidents \cite{Boyd_Access_2020}.

The post-mortem interval (PMI), while not allowing for direct person's identification, may play an important role in (a) narrowing down the time frame within which the individual likely died and was left at the scene (what directly facilitates biometric 1:N matching), and (b) conditioning iris recognition methods by the PMI to allow for different processing of the iris pattern depending on the severity of iris tissue decomposition, which is correlated with the PMI. This paper makes the first known to us attempt to assess the feasibility of the PMI estimation from post-mortem iris images, and offers methods for such iris image-based PMI assessment. More specifically, the {\bf novel contributions} are:

\begin{itemize}
    \setlength\itemsep{0em}
    \item[(a)] \textcolor{black}{deep learning-based 
    {\bf models estimating the PMI} from either near-infrared (NIR) ISO/IEC 19794-6-compliant or visible-light (RGB) iris images,}
    \item[(b)] a {\bf fusion model}, which combines NIR image- and RGB image-based predictions, if such multispectral forensic iris images are available,
    \item[(c)] {\bf very rigorous cross-validation experiments}, in which multiple models are independently trained and tested on 15,279 forensic iris images originating from 348 deceased subjects, in three scenarios:
    \begin{itemize}
        \setlength\itemsep{0em}
        \item[(S1)] sample-disjoint (but not subject-disjoint),
        \item[(S2)] subject-disjoint (but not dataset-disjoint), and
        \item[(S3)] dataset-disjoint (thus also subject-disjoint) with two data balancing approaches: up-sampling, with replacement, minority classes ({\it S3-real}), and adding synthetic post-mortem iris images to the training data to make the PMI distribution uniform ({\it S3-synthetic}).
    \end{itemize}
    \item[(d)] {\bf model weights and sources codes} to facilitate an immediate use of the designed models\footnote{The link to the repository has been removed to maintain anonymity, and will be added in the event of this paper being accepted.
    }.
\end{itemize}

The scenario (S3) simulates the most realistic situation when train and test data originate from different subjects, environmental conditions and acquisition procedures. In this challenging scenario, the best model utilizing the multispectral iris images (NIR+RGB) achieves the Mean Absolute Error (MAE) of 45.77 ($\pm$1.15) hours, or approx. 2 days. In a more favorable, but still subject-disjoint scenario (S2), in which the acquisition procedures are similar, the best model's MAE = 17.52 ($\pm17.19$) hours, so less than a day. Given that no other information than the iris image is available, these results are very encouraging and may contribute to a multi-modal PMI estimation if other type of data or information about the case are known.

%% file: sections/related-works.tex
\paragraph{Forensic Iris Recognition: }
For a considerable duration, the scientific and industrial communities have held the notion that iris recognition is challenging or even impossible to perform after a person's death. This belief is exemplified by Daugman's assertion in a 2001 BBC interview, where he highlighted that "shortly after death, the pupil dilates and the cornea becomes cloudy," rendering iris recognition problematic \cite{daugman2001bbc}. Similarly, Szczepanski et al. \cite{szczepanski2014pupil} claimed that "the iris decays within a few minutes after death," indicating a rapid degradation process. Moreover, certain commercial sources have asserted the scientific impossibility of utilizing a deceased individual's iris for recognition due to factors such as the relaxation of the iris muscle after death, resulting in a fully dilated pupil with no discernible iris texture. Consequently, it has been widely accepted that the lack of usable iris area \cite{IrisGuard2016} or the disappearance of iris features alongside pupil dilation \cite{IriTech2015} renders a deceased person's iris unsuitable for recognition purposes. However, recent investigations suggest that the decomposition of the eye and iris is more complex and slower than previously presumed. 

For instance, Bolme et al. \cite{bolme2016impact} were the pioneers in examining the post-mortem biometric performance of face, fingerprint, and iris recognition in outdoor environments. Their findings revealed that while fingerprints and faces showed moderate resilience to decomposition, iris recognition suffered rapid degradation. The correct verification rate dropped significantly, approaching zero after 14 days of outdoor exposure. A subsequent investigation by Sauerwein et al. \cite{sauerwein2017effect} demonstrated that irises can remain readable for up to 34 days post-mortem when cadavers are subjected to low temperatures during winter. Notably, human examiners were asked to match image pairs instead of using iris recognition algorithms. These findings suggest that exposing a cadaver to low temperatures in winter can increase the probability of correctly identifying an iris, even if it has been outside for an extended period.

A recent investigation by Trokielewicz et al. \cite{trokielewicz2018iris} suggests that post-mortem iris recognition may achieve close-to-perfect accuracy approximately 5–7 hours after death, and in some cases, remains viable even up to 21 days post-mortem. These findings challenge past assertions in the literature regarding the rapid degradation of the iris shortly after death, indicating that the dynamics of post-mortem changes to the iris, crucial for biometric identification, are more moderate than previously thought.

\paragraph{Post-mortem Interval Estimation:} Estimating the post-mortem interval remains one of the most needed and challenging tasks in forensic sciences. It typically relies on physical postmortem alterations like cooling \cite{kaliszan2009estimation, rodrigo2015time, igari2016rectal, rodrigo2016nonlinear}, stiffening \cite{varetto2005long, ozawa2013effect, nishida2015blood, martins2015necromechanics}, and decomposition \cite{vass2002decomposition, ferreira2013can, cockle2015human, suckling2016longitudinal}, as well as chemical changes such as electrolyte modifications in body fluids \cite{chandrakanth2013postmortem, cordeiro2015application, zilg2015new, parmar2015estimation, rognum2016estimation}. Forensic entomology has also been explored as a means to predict the PMI, involving the examination of insects' presence, age, and timing of insects' incidence on corpses \cite{tuccia2016combined, iancu2016dynamics}. Furthermore, various techniques utilizing signal and image processing methods have been suggested for PMI estimation. Canturk \etal \cite{canturk2016experimental} investigated the relationship between tissue conductivity changes and the time of death. Several studies have highlighted the potential of postmortem opacity development in the eye as a tool for estimating PMI. Kumar \etal \cite{kumar2012determination} provided insights into corneal opacity development based on personal observations, while Zhou \etal \cite{zhou2010image} and Liu \etal \cite{liu2008image} conducted studies on postmortem eye changes in rabbits, utilizing image processing methods for feature extraction and classification. Kawashima \etal \cite{kawashima2015estimating} investigated human subjects, employing RGB pixel values of corneal regions and developing a mathematical formula for the PMI calculation. Additionally, Canturk \etal \cite{canturk2017investigation} analyzed eye images from ten human subjects over a 15-hour period, medically interpreting postmortem alterations of the eye.

To our knowledge, {\bf no study has yet investigated the PMI estimation directly from  biometric iris images, and this paper proposes and evaluates the first such models}, being able to assess the PMI from either biometric (ISO/IEC 19794-6-compliant) or visible-light iris images.

%% file: sections/methodology.tex
\subsection{Datasets}
\label{sec:dataset}

To conduct this research, we have compiled \textcolor{black}{(from public sources)} a dataset of 8,064 NIR and 7,215 RGB forensic iris images derived from 348 subjects. At the time of preparing this paper, and to our knowledge, this is the largest dataset of forensic iris images that one can compile from publicly-available sources, which are characterized shortly below.

\textbf{Warsaw BioBase Post Mortem Iris v2.0} \cite{trokielewicz2018iris} includes 1,787 RGB and 1,200 NIR iris images sourced from 37 deceased individuals during from 1 to 13 acquisition sessions spanning 5 to 814 hours post-mortem. The data was collected in a hospital mortuary setting, and an ambient temperature was kept at approximately \ang{6} Celsius (\ang{42.8} Fahrenheit). Other details regarding pre-cold storage conditions, air pressure, and humidity, are unknown.

\textbf{Warsaw BioBase Post Mortem Iris v3.0} \cite{trokielewicz2020post}, comprises 785 RGB and 1,094 NIR images sourced from 42 deceased individuals, spanning up to 369 hours post-mortem. The environmental conditions mirror those known for the Warsaw BioBase Post Mortem Iris v2.0 dataset.

\textbf{NIJ-2018-DU-BX-0215} \cite{Czajka2023software}, stands as the most recent and extensive forensic iris dataset collected to date. It includes 4,643 RGB and 5,770 NIR images acquired from 269 deceased individuals, with PMI spanning up to 1,674 hours post-mortem. The acquisition was carried out during routine operation of the medical examiner in the Dutchess County Medical Examiner's Office, NY, USA.

\begin{figure*}[ht]
  \centering
  \includegraphics[width=\textwidth]{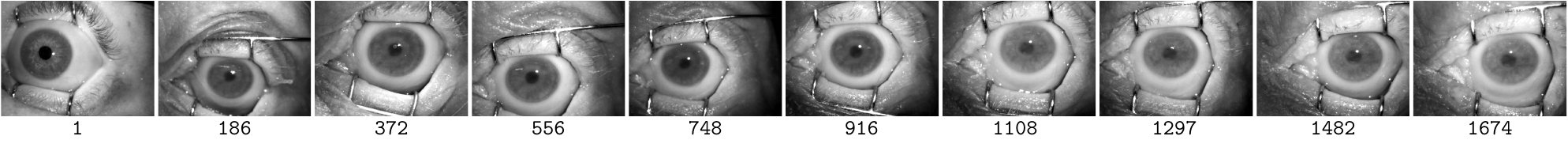} 
  \vspace{-0.8cm}
  \caption{Example post-mortem iris samples collected at different post-mortem intervals (shown underneath the images).}
  \label{fig:sample}
\end{figure*}

\textcolor{black}{All datasets offer the metadata: gender, age, and PMI, and Warsaw subsets also offer the information about the death reason.} We will call all samples from the {\bf Warsaw BioBase Post Mortem Iris v2.0} and {\bf Warsaw BioBase Post Mortem Iris v3.0} datasets combined as {\bf Warsaw}, since all these samples were collected in the same environment. Also for the sake of brevity, we will call {\bf NIJ-2018-DU-BX-0215} as {\bf NIJ}.

Fig. \ref{fig:sample} illustrates how iris image (and iris pattern specifically) changes when PMI grows. It is also important to note that the PMI distributions in Warsaw and NIJ datasets are different. It is evident from Fig. \ref{fig:dataset_box_plot} that the NIJ dataset includes samples from a more diverse range of the PMI.

\input{figure_tex/dataset-disjoint-box-plot}

\subsection{Data Balancing}
\label{sec:data_balancing}
For scenario S3, the training data is balanced through two methods: up-sampling with real data ({\it S3-real}) and supplementing the real iris scans with synthetic samples generated by a conditional StyleGAN2-ADA ({\it S3-synthetic}). 

To implement {\it S3-real}, we divided the training set into 18 PMI ranges, as proposed in \cite{Bhuiyan_WACVW_2024} (class 1 representing images with PMI of 0-24h, class 2: PMI of 25h-48h, and so on, until class 18 representing images with PMI larger than 409h). We identified the class with the maximum number of samples and up-sampled (with replacement) the other classes to match this number for both NIR and RGB data. 

To implement {\it S3-synthetic}, and instead of up-sampling with real data, we supplemented each class to ensure an equal number of samples across all classes with synthetically-generated forensic iris images. For NIR domain, we used synthetic samples offered in \cite{Bhuiyan_WACVW_2024}. For RGB domain, we trained a separate StyleGAN2-ADA and synthesized 3,000 RGB samples for each PMI class. The exact PMI values were randomly assigned to each sample within their respective class PMI range.

\subsection{PMI Estimation Models}

\textcolor{black}{We explore several models with diverse architectures to build regression models estimating the PMI, including four convolutional neural networks (CNN) backbones: VGG19 \cite{simonyan2014very} (classical CNN), DenseNet121 \cite{huang2017densely} (incorporating dense connections), ResNet152 \cite{he2016deep} (incorporating residual connections), and Inception\_v3 \cite{szegedy2016rethinking} (implementing inception modules). Additionally for the most challenging scenario {\bf S3}, we used a Vision Transformer (ViT) \cite{dosovitskiy2020image} and a domain-specific ResNet152 (DS-ResNet152) pre-trained for iris presentation attack detection.} Furthermore, to discern the most effective strategy for PMI estimation, the proposed models utilize (1) only NIR samples, (2) only RGB samples, and (3) a multispectral approach in which both NIR and RGB samples are used (assuming their availability from the same subjects). These three approaches are characterized briefly below.

In our narrow-band spectral models (NIR and RGB, considered individually), we modified the last (classification) layer for the models fed with RGB images, while for the models utilizing NIR samples, we modified both the first and last layers of each backbone. This was needed to adapt the model backbones to the regression task, and to different number of channels in NIR and RGB images.

Our multispectral models (NIR and RGB samples combined) consist of two separate models: $\mathbf{f}_{\text{NIR}}$ dedicated to processing NIR data and $\mathbf{f}_{\text{RGB}}$ processing the RGB data. Both models employ the same model architectures used in narrow-band spectral models, ensuring a consistent approach across different modalities. The PMI prediction $\hat{y}$ is made by a simple two-layer perceptron, put on top of concatenated embeddings extracted by $\mathbf{f}_{\text{NIR}}$ and $\mathbf{f}_{\text{RGB}}$, namely:
\vskip-2mm
\begin{equation}
\hat{y} = \mathbf{W}_2 \sigma\big(\mathbf{W}_1 (\mathbf{f}_{\text{NIR}} \oplus \mathbf{f}_{\text{RGB}}) + \mathbf{b}_1\big) + b_2
\end{equation}

\noindent
where $\oplus$ is a vector concatenation operator, $\mathbf{W}_1$ and $\mathbf{W}_2$ are weight matrix and vector, $\mathbf{b}_1$ is the bias vector, $b_2$ is a scalar bias term, and $(\sigma)$ is a ReLU activation function.

\subsection{Data Preprocessing and Augmentation}

\begin{figure}
    \centering
    \includegraphics[width=0.6\linewidth]{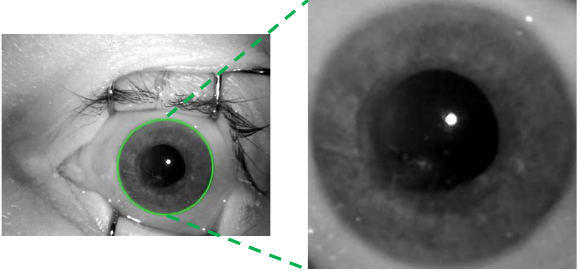}
    \vspace{-0.2cm}
    \caption{Original image (left) and cropped region used as model input (right).}
    \label{fig:sample-input}
\end{figure}

\textcolor{black}{In this study, we first locate the iris outer boundary and crop it (see Fig. \ref{fig:sample-input}) to $299\times299$ pixel square for the Inception\_v3 backbone, and to $224\times224$ pixel square for other backbones.} There are two reasons of using cropped images. First, the periocular regions of post-mortem irises include features, such as metal retractors used to increase the palpebral fissure, that may be accidentally correlated with the PMI, and picked by the models. Secondly, we were interested in applying pre-trained models due to sparse and small datasets of post-mortem iris images that the community has at hand right now, what makes training the models from scratch more challenging. \textcolor{black}{Initially, we did experiments with normalized iris images, but observed significantly worse results compared to using ISO-compliant cropped images, so we proceeded with the latter in this study.} Standard augmentations were applied to the training samples, such as random horizontal flipping, random rotation within a range of -30 to 30 degrees, random brightness, contrast, and sharpness adjustments.

\subsection{Performance Evaluation and Metrics}
\label{subsec:perf-eval}

We evaluate our models using three distinct scenarios: {\bf (S1) sample-disjoint} 10-fold cross-validation, {\bf (S2) subject-disjoint} 10-fold cross-validation, and {\bf (S3) cross-dataset} evaluation. To ensure a fair comparison, we fixed the training and testing datasets across all models during experimentation. 

In all training experiments, we used Adam optimizer with a batch size of 32, a learning rate of 1e-4, and a weight decay of 1e-6. We trained each model for 500 epochs.

To assess the effectiveness of the proposed PMI estimation model, we calculate Root-Mean Square Error (RMSE) and Mean Absolute Error (MAE) between the ground truth PMI and predicted PMI.

%% file: figure_tex/dataset-disjoint-box-plot.tex
\begin{figure}[!ht] 
    \centering
    \begin{subfigure}[b]{0.48\linewidth}
        \centering
        \includegraphics[width=\linewidth]{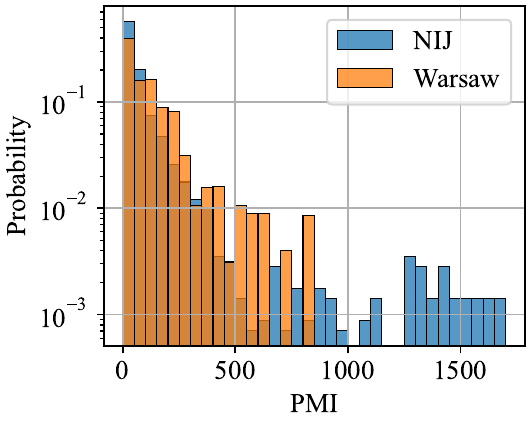}
        \caption{NIR samples only}
        \label{dataset:nir}
    \end{subfigure}%
    \begin{subfigure}[b]{0.48\linewidth}
        \centering
        \includegraphics[width=\linewidth]{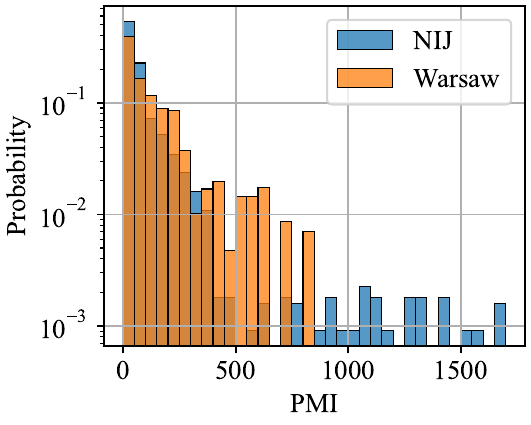}
        \caption{RGB samples only}
        \label{dataset:rgb}
    \end{subfigure}%
    \\
    \begin{subfigure}[b]{0.48\linewidth}
        \centering
        \includegraphics[width=\linewidth]{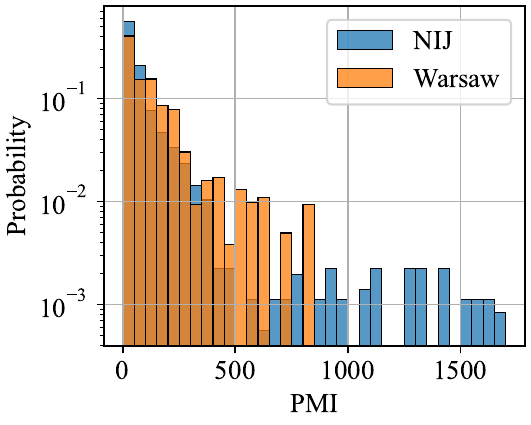}
        \caption{Multispectral}
        \label{dataset:multispectral}
    \end{subfigure}%
    \vspace{-0.2cm}
    \caption{PMI distributions for the Warsaw and NIJ datasets for NIR-only samples, RGB-only samples, and NIR and RGB samples (pairs taken from the same eye) combined.}
    \label{fig:dataset_box_plot}
\end{figure}

%% file: sections/experiment.tex
\subsection{Sample-disjoint Evaluation (S1)}

In this scenario, we assess the performance of our models using a sample-disjoint 10-fold cross-validation approach. This is the most favorable scenario, in which we assume that (a) previous data, although with different PMIs, from the same subjects is available, and (b) the acquisition environment (\eg, sensors, ambient temperature, technicians collecting the data) is the same in case of training and test samples. This scenario S1 sets the lower bound for the errors in estimating the PMI solely from the iris image.

\input{table_tex/10f_nir_table}
\input{table_tex/10f_rgb_table}
\input{table_tex/10f_multispectral_table}

Tables \ref{tab:nir}, \ref{tab:rgb}, and \ref{tab:multispectral} provide a comparative analysis of the PMI estimation model's average performance, measured in hours, across different model backbones for NIR-only, RGB-only, and multispectral data, respectively. For the NIR data (Tab. \ref{tab:nir}), the DenseNet121 backbone achieves the lowest RMSE (\textbf{7.18 hours}) and MAE (\textbf{4.10 hours}) among the models, indicating superior performance in estimating the PMI for NIR spectral data. Conversely, the Inception\_v3 backbone exhibits significantly higher RMSE (\textbf{92.00 hours}) and MAE (\textbf{56.35 hours}) compared to other models, suggesting less accurate PMI estimates for NIR data. 

Similarly, for RGB data (Tab. \ref{tab:rgb}), the DenseNet121 backbone again demonstrates the lowest RMSE (\textbf{14.32 hours}) and MAE (\textbf{5.75 hours}), while Inception\_v3 once more exhibits higher RMSE (\textbf{69.32 hours}) and MAE (\textbf{38.43 hours}). 

Finally, in the case of multispectral data (Tab. \ref{tab:multispectral}), the ResNet152 backbone achieves the lowest RMSE (\textbf{6.18 hours}) and MAE (\textbf{3.44 hours}) while the Inception\_v3 backbone performing better compared to its performance with NIR and RGB data individually, still exhibits higher RMSE (\textbf{38.59 hours}) and MAE (\textbf{20.37 hours}). 

\input{figure_tex/10f_multispectral_model_plot}

In Figure \ref{fig:multispectral_model_plot}, we observe distinct patterns among the backbone models. Specifically, the VGG19, DenseNet121, and ResNet152 models demonstrate the accurate estimation of higher PMI. However, these models exhibit a few discrepancies in their estimations for lower PMI and the differences between the predicted and actual PMI are consistently small within the entire PMI range. The Inception\_v3 model more often overestimates and underestimates the PMI specially for a higher PMI range. Similar scatter plots for NIR and RGB data are included in the Supplementary Materials (Figures \ref{fig:nir_model_plot} and \ref{fig:RGB_model_plot}).

\input{figure_tex/10f_multispectral_box_plot}

This spectacularly good performance in PMI estimation (with errors for the best model not exceeding 4 hours) is due to two factors. First, as mentioned above, is the non-subject-disjoint train-test protocol. Second, as demonstrated in Fig.  \ref{fig:multispectral_10f_box_plot}, is the wider range of the PMI in the training set compared to the test set, for both the worst model and the best model obtained in 10-fold train-test cross-validation. Figures \ref{fig:nir_10f_box_plot} and \ref{fig:multispectral_10f_box_plot} for NIR and RGB data, separately, show very similar trends, and where thus included in the Supplementary Materials. 

\subsection{Subject-disjoint Evaluation (S2)}

In this scenario, we examine the performance of our models under a subject-disjoint 10-fold evaluation setup. This evaluation approach involves partitioning the combined datasets (Warsaw and NIJ) so that samples from each subject are exclusively present in either the training or the test set, what makes it more realistic than scenario S1. This scenario assumes, as in S1, that acquisition environment is uniform across the train and test data collections. 

\input{table_tex/nir_subject_disjoint_10f}
\input{table_tex/rgb_subject_disjoint_10f}
\input{table_tex/multispectral_subject_disjoint_10f}

Tables \ref{tab:nir-subject-disjoint}, \ref{tab:rgb-subject-disjoint}, and \ref{tab:multispectral-subject-disjoint} compare the PMIs estimated by all model architectures for NIR-only, RGB-only, and multispectral data, respectively. For the NIR data (Tab. \ref{tab:nir-subject-disjoint}), the DenseNet121 backbone achieves the lowest RMSE (\textbf{31.23 hours}) and MAE (\textbf{19.67 hours}) among the models. Conversely, the Inception\_v3 backbone exhibits significantly higher RMSE (\textbf{115.07 hours}) and MAE (\textbf{68.79 hours}) compared to other models, suggesting less accurate PMI estimates for NIR data. 

Similarly, for RGB data (Tab. \ref{tab:rgb-subject-disjoint}), the VGG19 backbone demonstrates the lowest RMSE (\textbf{47.96 hours}) and MAE (\textbf{27.20 hours}), while Inception\_v3 backbone once more exhibits higher RMSE (\textbf{77.74 hours}) and MAE (\textbf{45.90 hours}).

\input{figure_tex/subject_disjoint_10f_multispectral_model_plot}

In the case of multispectral data (Tab. \ref{tab:multispectral-subject-disjoint}), the VGG19 backbone achieves the lowest RMSE (\textbf{29.00 hours}) and MAE (\textbf{17.52 hours}) while the Inception\_v3 backbone performing better compared to its performance with NIR and RGB data individually, still exhibits higher RMSE (\textbf{56.33 hours}) and MAE (\textbf{32.42 hours}). The trends observed in Figure \ref{fig:subject-disjoint-multispectral} are similar across almost all model backbones, which tend to underestimate the PMI, especially for higher actual PMI values. Inception\_v3, however, occasionally overestimates the PMI for higher PMI ranges. One potential explanation is the scarcity of data with PMIs larger than 1000 hours, making it difficult for the models to learn features associate with iris pattern decomposition for such high PMIs. These observations are very similar to models trained solely on NIR or RGB data, and thus the scatter plots for these cases were included in the supplementary materials (Figures \ref{fig:subject-disjoint-nir} and \ref{fig:subject-disjoint-rgb}).

\input{figure_tex/subject_disjoint_10f_multispectral_box_plot}

Fig. \ref{fig:multispectral_subject_box_plot} illustrates how PMI distributions in training and test datasets differ when we look at folds in which the best or the worst results were obtained (for multispectral data). The least-performing fold included test samples with a wider PMI range compared to train samples, while the best-performing fold has an opposite relationship. \textcolor{black}{This suggests that a large variation of the PMI in the training set may be a factor in achieving good performance.} An identical observation has been made for models trained with the NIR and RGB data, separately (corresponding Figs \ref{fig:nir_subject_box_plot} and \ref{fig:rgb_subject_box_plot} are added to the supplementary materials).

\subsection{Cross-dataset Evaluation (S3)}
In this scenario, we evaluate the performance of the models in a training regime that is both subject-disjoint {\bf and} dataset-disjoint (\ie, data has been collected in two different laboratories by two different teams). This scenario thus does not assume knowledge of the acquisition environment, making it the most realistic of all the scenarios considered in this work. As described in Sec. \ref{sec:data_balancing} we explore using existing data (no data balancing) and two training data balancing strategies: up-sampling (with replacement) from real data ({\it S3-real}) and up-sampling from synthetically-generated data ({\it S3-synthetic}) to make the PMI distribution in the training set uniform.

Tab. \ref{tab:testset-warsaw-nir-data} compares models trained on the NIJ dataset and tested on the Warsaw dataset using NIR data. Among these models, Inception\_v3 with {\it S3-synthetic} balancing achieves the lowest RMSE (\textbf{106.29 hours}) and MAE (\textbf{65.61 hours}). Tab. \ref{tab:testset-nij-nir-data} compares models after swapping the datasets, \ie, models are trained on Warsaw and tested on NIJ dataset, where Inception\_v3 demonstrates the lowest RMSE (\textbf{117.31 hours}) and MAE (\textbf{55.84 hours}) with {\it S3-synthetic} balancing, showcasing consistent performance across datasets.

Tab. \ref{tab:testset-warsaw-rgb-data} illustrates the results for models trained with RGB data. Initially, models are trained on the NIJ dataset and tested on the Warsaw dataset, where Densenet121 stands out with the lowest RMSE (\textbf{70.07 hours}) and MAE (\textbf{48.54 hours}). After swapping the datasets, models are trained on the Warsaw dataset and tested on the NIJ dataset (Tab. \ref{tab:testset-nij-rgb-data}). In this scenario, Inception\_v3 demonstrates the lowest RMSE (\textbf{83.83 hours}) and MAE (\textbf{54.94 hours}). In both cases for RGB data, the synthetic balancing technique improved the model's performance significantly. Also, compared to the results obtained for NIR samples, RGB images, by exposing more decomposition-related deformations than NIR samples, offer a better PMI prediction.

Finally, Tabs. \ref{tab:testset-warsaw-multi-data} and \ref{tab:testset-nij-multi-data} illustrate analogous experiments using multispectral data. In these cases, Inception\_v3 and Resnet152 achieve the best accuracy with {\it S3-synthetic} balancing technique, achieving the lowest RMSE (\textbf{89.55 hours} and \textbf{92.18 hours}, respectively) and MAE (\textbf{54.99 hours} and \textbf{45.77 hours}, respectively) across datasets. It is thus very interesting to see that supplementing real and heavily unbalanced training data with appropriately-synthesized (\ie simulating a given PMI) samples significantly improved the model's ability to generalize to new data collected in a different forensic laboratory and by a different team.

\input{new_table_tex/nir-testset-warsaw}
\input{new_table_tex/rgb-testset-warsaw}
\input{new_table_tex/multi-testset-warsaw}

\input{new_table_tex/nir-testset-nij}
\input{new_table_tex/rgb-testset-nij}
\input{new_table_tex/multi-testset-nij}

\input{new_figure_tex/war-multi-best-model}
\input{new_figure_tex/nij-multi-best-model}

%% file: table_tex/10f_nir_table.tex
\begin{table}[!ht]
\footnotesize
\centering
\caption{PMI (in hours) estimated on the \textbf{NIR data} and averaged across \textbf{sample-disjoint} 10-fold cross-validation experiments.}
\vspace{-0.3cm}
\label{tab:nir}
\vskip1mm
\begin{tabular}{lcccc}
\toprule
 & \multicolumn{2}{c}{RMSE} & \multicolumn{2}{c}{MAE} \\ \cmidrule(lr){2-3} \cmidrule(lr){4-5}
\multirow{-2}{*}{\begin{tabular}[c]{@{}l@{}}Model \\ Backbone\end{tabular}} & Mean & StDev & Mean & StDev \\ \midrule
VGG19         & 9.31      & 8.32      & 5.04     & 2.62      \\
Inception\_v3 & 92.00     & 14.07     & 56.35    & 7.78      \\
\textbf{DenseNet121}   & \textbf{7.18}      & \textbf{7.14}      & \textbf{4.10}     & \textbf{2.74}      \\
ResNet152     & 7.52      & 7.15      & 4.20     & 2.12      \\
\bottomrule
\end{tabular}
\end{table}

%% file: table_tex/10f_rgb_table.tex
\begin{table}[ht]
\footnotesize
\centering
\caption{Same as in Tab. \ref{tab:nir} for \textbf{RGB data}.}
\vspace{-0.3cm}
\label{tab:rgb}
\vskip1mm
\begin{tabular}{lcccc}
\toprule
 & \multicolumn{2}{c}{RMSE} & \multicolumn{2}{c}{MAE} \\ \cmidrule(lr){2-3} \cmidrule(lr){4-5}
\multirow{-2}{*}{\begin{tabular}[c]{@{}l@{}}Model \\ Backbone\end{tabular}} & Mean & StDev & Mean & StDev \\ \midrule
VGG19         & 18.09     & 11.87     & 9.24     & 3.34     \\
Inception\_v3 & 69.32     & 18.98     & 38.43    & 4.09     \\
\textbf{DenseNet121}   & \textbf{14.32}     & \textbf{18.01}     & \textbf{5.75}     & \textbf{3.39}     \\
ResNet152     & 19.29     & 11.66     & 5.77     & 3.30     \\
\bottomrule
\end{tabular}
\end{table}

%% file: table_tex/10f_multispectral_table.tex
\begin{table}[ht]
\footnotesize
\centering
\caption{Same as in Tab. \ref{tab:nir} for \textbf{Multispectral data}.}
\vspace{-0.3cm}
\label{tab:multispectral}
\begin{tabular}{lcccc}
\toprule
 & \multicolumn{2}{c}{RMSE} & \multicolumn{2}{c}{MAE} \\ \cmidrule(lr){2-3} \cmidrule(lr){4-5}
\multirow{-2}{*}{\begin{tabular}[c]{@{}l@{}}Model \\ Backbone\end{tabular}} & Mean & StDev & Mean & StDev. \\ \midrule
VGG19         & 9.32      & 8.11      & 5.40     & 3.67     \\
Inception\_v3 & 38.59     & 4.32      & 20.37    & 1.91     \\
DenseNet121   & 6.16      & 7.01      & 3.56     & 3.09     \\
\textbf{ResNet152}     & \textbf{6.18}      & \textbf{4.79}      & \textbf{3.44}     & \textbf{2.28}     \\
\bottomrule
\end{tabular}
\end{table}

%% file: figure_tex/10f_multispectral_model_plot.tex
\begin{figure*}[ht] 
    \centering
    \begin{subfigure}[b]{0.24\linewidth}
        \centering
        \includegraphics[width=\linewidth]{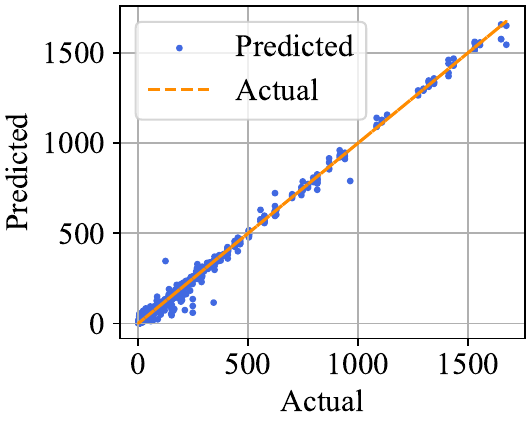}
        \caption{VGG19}
        \label{multispectral:vgg19}
    \end{subfigure}%
    \begin{subfigure}[b]{0.24\linewidth}
        \centering
        \includegraphics[width=\linewidth]{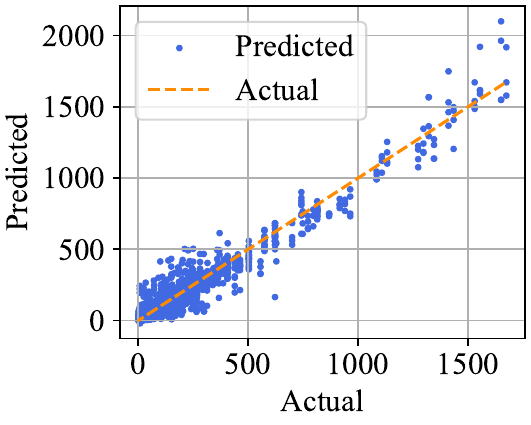}
        \caption{Inception\_v3}
        \label{multispectral:inception}
    \end{subfigure}%
    \begin{subfigure}[b]{0.24\linewidth}
        \centering
        \includegraphics[width=\linewidth]{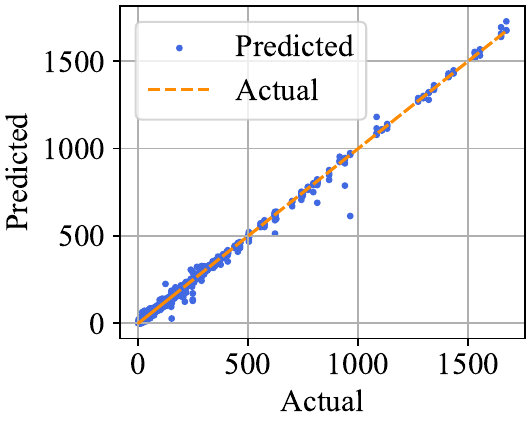}
        \caption{DenseNet121}
        \label{multispectral:densenet121}
    \end{subfigure}%
    \begin{subfigure}[b]{0.24\linewidth}
        \centering
        \includegraphics[width=\linewidth]{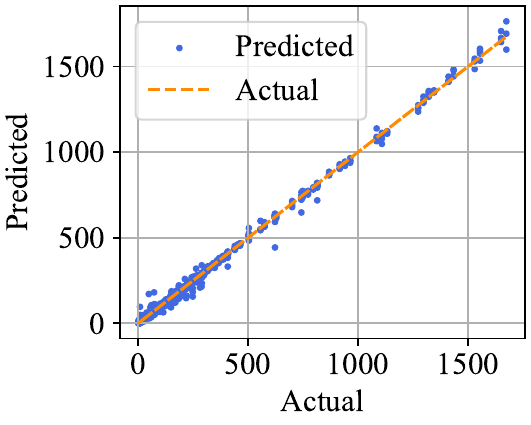}
        \caption{ResNet152}
        \label{multispectral:resnet152}
    \end{subfigure}
    \vspace{-0.2cm}
    \caption{Scatter plots visualizing the predicted PMI values against the actual PMI values for \textbf{multispectral data} combined for all \textbf{sample-disjoint} 10-fold cross-validations.}
    \label{fig:multispectral_model_plot}
\end{figure*}

%% file: figure_tex/10f_multispectral_box_plot.tex
\begin{figure}[ht] 
    \centering
    \begin{subfigure}[b]{0.48\linewidth}
        \centering
        \includegraphics[width=\linewidth]{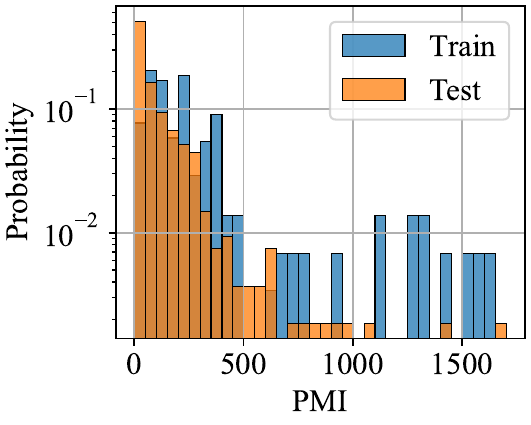}
        \caption{Least performance fold}
        \label{10f-multispectral:least-fold}
    \end{subfigure}%
    \begin{subfigure}[b]{0.48\linewidth}
        \centering
        \includegraphics[width=\linewidth]{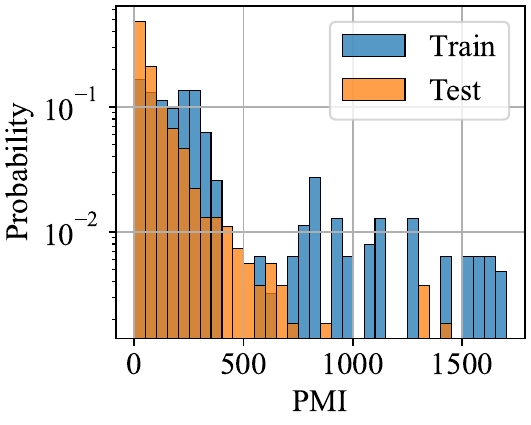}
        \caption{Best performance fold}
        \label{10f-multispectral:best-fold}
    \end{subfigure}%
    \vspace{-0.2cm}
    \caption{Comparison of the PMI distributions between training and test datasets for the least (a) and best (b) performing folds observed in the \textbf{sample-disjoint} 10-fold cross-validation for the \textbf{multispectral data}.}
    \label{fig:multispectral_10f_box_plot} 
\end{figure} 

%% file: table_tex/nir_subject_disjoint_10f.tex
\begin{table}[ht]
\footnotesize
\centering
\caption{PMI (in hours) estimated on the \textbf{NIR data} and averaged across \textbf{subject-disjoint} 10-fold cross-validation experiments.}
\vspace{-0.3cm}
\label{tab:nir-subject-disjoint}
\vskip1mm
\begin{tabular}{lcccc}
\toprule
 & \multicolumn{2}{c}{RMSE} & \multicolumn{2}{c}{MAE} \\ \cmidrule(lr){2-3} \cmidrule(lr){4-5}
\multirow{-2}{*}{\begin{tabular}[c]{@{}l@{}}Model \\ Backbone\end{tabular}} & Mean & StDev & Mean & StDev \\ \midrule
VGG19         & 55.54     & 56.85     & 32.35    & 24.72    \\
Inception\_v3 & 115.07    & 67.46     & 68.79    & 31.95    \\
\textbf{DenseNet121}   & \textbf{31.23}     & \textbf{30.20}     & \textbf{19.67}    & \textbf{16.14}    \\
ResNet152     & 45.53     & 47.68     & 26.49    & 21.98    \\
\bottomrule
\end{tabular}
\end{table}

%% file: table_tex/rgb_subject_disjoint_10f.tex
\begin{table}[ht]
\footnotesize
\centering
\caption{Same as in Tab. \ref{tab:nir-subject-disjoint} for \textbf{RGB data}.}
\vspace{-0.3cm}
\label{tab:rgb-subject-disjoint}
\vskip1mm
\begin{tabular}{lcccc}
\toprule
 & \multicolumn{2}{c}{RMSE} & \multicolumn{2}{c}{MAE} \\ \cmidrule(lr){2-3} \cmidrule(lr){4-5}
\multirow{-2}{*}{\begin{tabular}[c]{@{}l@{}}Model \\ Backbone\end{tabular}} & Mean & StDev & Mean & StDev \\ \midrule
\textbf{VGG19}         & \textbf{47.96}     & \textbf{35.35}     & \textbf{27.20}    & \textbf{15.72}    \\
Inception\_v3 & 77.74     & 40.78     & 45.90    & 19.75    \\
DenseNet121   & 50.53     & 28.13     & 33.90    & 16.64    \\
ResNet152     & 52.14     & 46.98     & 30.91    & 21.36    \\
\bottomrule
\end{tabular}
\end{table}

%% file: table_tex/multispectral_subject_disjoint_10f.tex
\begin{table}[ht]
\footnotesize
\centering
\caption{Same as in Tab. \ref{tab:nir-subject-disjoint} for \textbf{multispectral data}.}
\vspace{-0.3cm}
\label{tab:multispectral-subject-disjoint}
\begin{tabular}{lcccc}
\toprule
 & \multicolumn{2}{c}{RMSE} & \multicolumn{2}{c}{MAE} \\ \cmidrule(lr){2-3} \cmidrule(lr){4-5}
\multirow{-2}{*}{\begin{tabular}[c]{@{}l@{}}Model \\ Backbone\end{tabular}} & Mean & StDev & Mean & StDev \\ \midrule
\textbf{VGG19}         & \textbf{29.00}     & \textbf{29.71}     & \textbf{17.52    }& \textbf{17.19}    \\
Inception\_v3 & 56.33     & 22.75     & 32.42    & 12.93    \\
DenseNet121   & 31.86     & 28.74     & 19.69    & 14.98    \\
ResNet152     & 30.72     & 28.62     & 20.02    & 19.01    \\
\bottomrule
\end{tabular}
\end{table}

%% file: figure_tex/subject_disjoint_10f_multispectral_model_plot.tex
\begin{figure*}[ht] 
    \centering
    \begin{subfigure}[b]{0.24\linewidth}
        \centering
        \includegraphics[width=\linewidth]{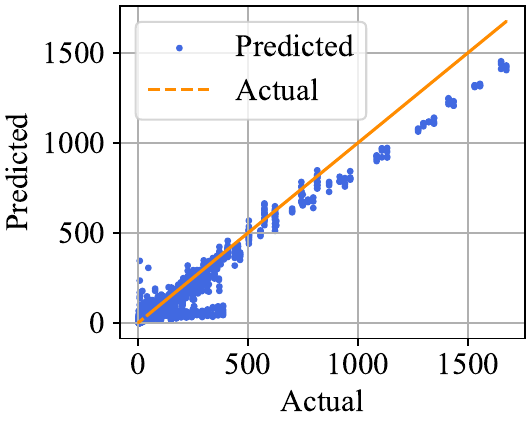}
        \caption{VGG19}
        \label{subject-multispectral:vgg19}
    \end{subfigure}%
    \begin{subfigure}[b]{0.24\linewidth}
        \centering
        \includegraphics[width=\linewidth]{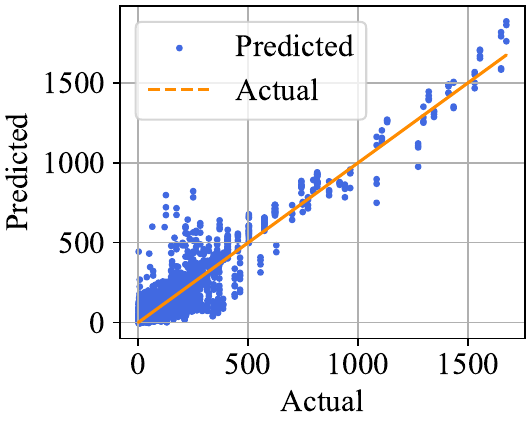}
        \caption{Inception\_v3}
        \label{subject-multispectral:inception}
    \end{subfigure}%
    \begin{subfigure}[b]{0.24\linewidth}
        \centering
        \includegraphics[width=\linewidth]{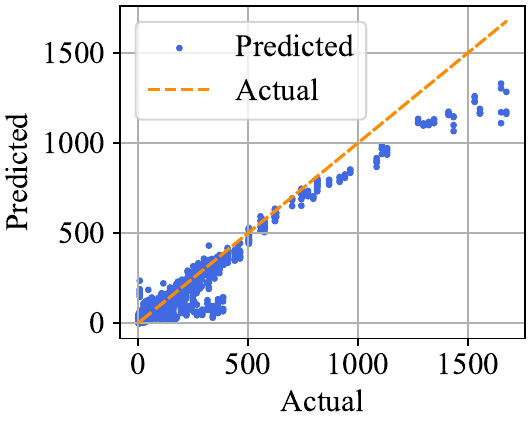}
        \caption{DenseNet121}
        \label{subject-multispectral:densenet121}
    \end{subfigure}%
    \begin{subfigure}[b]{0.24\linewidth}
        \centering
        \includegraphics[width=\linewidth]{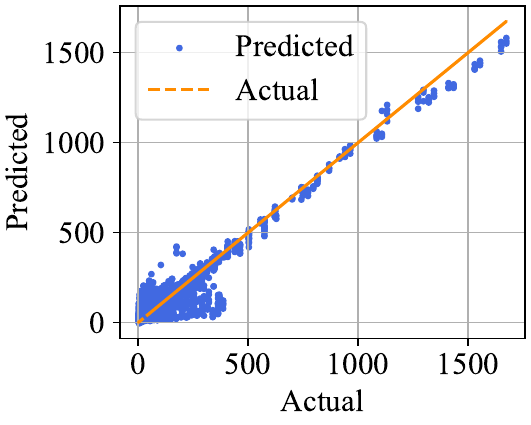}
        \caption{ResNet152}
        \label{subject-multispectral:resnet152}
    \end{subfigure}
    \vspace{-0.2cm}
    \caption{Scatter plots visualizing the predicted PMI values against the actual PMI values for \textbf{multispectral data} combined for all \textbf{subject-disjoint} 10-fold cross-validations.}
    \label{fig:subject-disjoint-multispectral} 
\end{figure*}

%% file: figure_tex/subject_disjoint_10f_multispectral_box_plot.tex
\begin{figure}[ht] 
    \centering
    \begin{subfigure}[b]{0.48\linewidth}
        \centering
        \includegraphics[width=\linewidth]{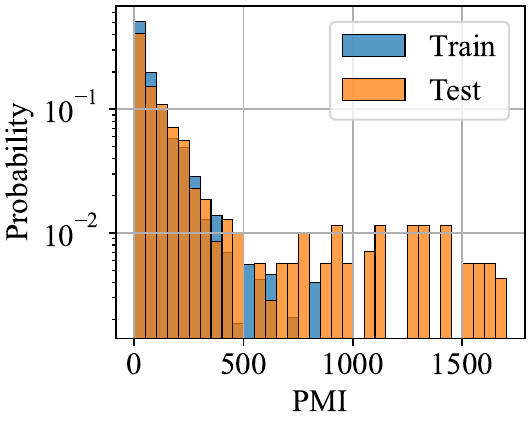}
        \caption{Least performance fold}
        \label{subject-multispectral:least-fold}
    \end{subfigure}%
    \begin{subfigure}[b]{0.48\linewidth}
        \centering
        \includegraphics[width=\linewidth]{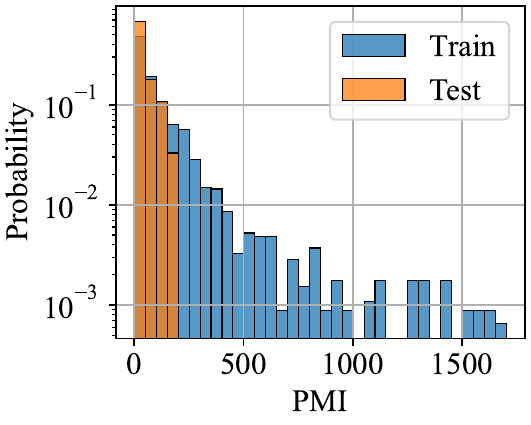}
        \caption{Best performance fold}
        \label{subject-multispectral:best-fold}
    \end{subfigure}%
    \vspace{-0.2cm}
    \caption{Comparison of the PMI distributions between training and test datasets for the least (a) and best (b) performing folds observed in the \textbf{subject-disjoint} 10-fold cross-validation for the \textbf{multispectral data}.}
    \label{fig:multispectral_subject_box_plot} 
\end{figure}

%% file: new_table_tex/nir-testset-warsaw.tex
\begin{table}[ht]
\caption{PMI (in hours) estimated on the \textbf{NIR data} by models trained on NIJ dataset and \textbf{tested on Warsaw} dataset.}
\vspace{-0.3cm}
\footnotesize
\label{tab:testset-warsaw-nir-data}
\begin{tabular}{@{}lcccccc@{}}
\toprule
\multirow{2}{*}{\begin{tabular}[c]{@{}l@{}}Model\\ Backbone\end{tabular}} & \multicolumn{2}{c}{\begin{tabular}[c]{@{}c@{}}Without \\ balancing\end{tabular}} & \multicolumn{2}{c}{\begin{tabular}[c]{@{}c@{}}{\it S3-real}\\ balancing\end{tabular}} & \multicolumn{2}{c}{\begin{tabular}[c]{@{}c@{}}{\it S3-synthetic}\\ balancing\end{tabular}} \\ \cmidrule(l){2-3} \cmidrule(l){4-5} \cmidrule(l){6-7}
 & RMSE & MAE & RMSE & MAE & RMSE & MAE \\ \cmidrule(r){1-1}
VGG19 & 112.71 & 73.14 & 119.85 & 76.76 & 107.32 & 68.1 \\
Inception\_v3 & 118.18 & 78.76 & 123.58 & 75.97 & \textbf{106.29} & \textbf{65.61} \\
DenseNet121 & 112.2 & 72.36 & 121.15 & 75.48 & 107.5 & 65.64 \\
ResNet152 & 105.62 & 71.48 & 125.95 & 77.89 & 110.07 & 65.89 \\ 
ViT & 147.35 & 90.25 & 153.10 & 95.76 & 153.83 & 95.76 \\ 
DS-ResNet152 & 113.38 & 84.80 & 119.78 & 75.59 & 105.18 & 66.66 \\
\bottomrule
\end{tabular}
\end{table}

%% file: new_table_tex/rgb-testset-warsaw.tex
\begin{table}[ht]
\caption{Same as in Tab. \ref{tab:testset-warsaw-nir-data} for \textbf{RGB data}.}
\vspace{-0.3cm}
\footnotesize
\label{tab:testset-warsaw-rgb-data}
\begin{tabular}{@{}lcccccc@{}}
\toprule
\multirow{2}{*}{\begin{tabular}[c]{@{}l@{}}Model\\ Backbone\end{tabular}} & \multicolumn{2}{c}{\begin{tabular}[c]{@{}c@{}}Without \\ balancing\end{tabular}} & \multicolumn{2}{c}{\begin{tabular}[c]{@{}c@{}}{\it S3-real}\\ balancing\end{tabular}} & \multicolumn{2}{c}{\begin{tabular}[c]{@{}c@{}}{\it S3-synthetic}\\ balancing\end{tabular}} \\ \cmidrule(l){2-3} \cmidrule(l){4-5} \cmidrule(l){6-7}
 & RMSE & MAE & RMSE & MAE & RMSE & MAE \\ \cmidrule(r){1-1}
VGG19 & 106.07 & 88.58 & 118.86 & 86.71 & 80.97 & 55.31 \\
Inception\_v3 & 90.14 & 73.23 & 115.75 & 85.34 & 71.57 & 49.09 \\
DenseNet121 & 99.68 & 82.32 & 110.15 & 81.26 & \textbf{70.07} & \textbf{48.54} \\
ResNet152 & 98.18 & 81.15 & 115.2 & 81.27 & 74.67 & 48.79 \\ 
ViT & 117.93 & 88.09 & 114.10 & 87.81 & 99.89 & 74.13 \\
DS-ResNet152 & 113.38 & 84.80 & 107.88 & 81.21 & 71.21 & 49.61 \\
\bottomrule
\end{tabular}
\end{table}

%% file: new_table_tex/multi-testset-warsaw.tex
\begin{table}[ht]
\caption{Same as in Tab. \ref{tab:testset-warsaw-nir-data} for \textbf{Multispectral data}.}
\vspace{-0.3cm}
\footnotesize
\label{tab:testset-warsaw-multi-data}
\begin{tabular}{@{}lcccccc@{}}
\toprule
\multirow{2}{*}{\begin{tabular}[c]{@{}l@{}}Model\\ Backbone\end{tabular}} & \multicolumn{2}{c}{\begin{tabular}[c]{@{}c@{}}Without \\ balancing\end{tabular}} & \multicolumn{2}{c}{\begin{tabular}[c]{@{}c@{}}{\it S3-real}\\ balancing\end{tabular}} & \multicolumn{2}{c}{\begin{tabular}[c]{@{}c@{}}{\it S3-synthetic}\\ balancing\end{tabular}} \\ \cmidrule(l){2-3} \cmidrule(l){4-5} \cmidrule(l){6-7}
 & RMSE & MAE & RMSE & MAE & RMSE & MAE \\ \cmidrule(r){1-1}
VGG19 & 107.68 & 69.58 & 129.12 & 77.95 & 91.69 & 56.63 \\
Inception\_v3 & 109.82 & 69.22 & 121.31 & 72.65 & \textbf{89.55} & \textbf{54.99} \\
DenseNet121 & 112.74 & 70.04 & 119.67 & 76.14 & 89.85 & 57.53 \\
ResNet152 & 111.31 & 69.12 & 122.95 & 72.84 & 105.34 & 66.21 \\ 
ViT & 142.31 & 85.54 & 139.95 & 93.49 & 93.97 & 64.18 \\
DS-ResNet152 & 116.80 & 74.21 & 117.47 & 71.49 & 107.33 & 67.83 \\
\bottomrule
\end{tabular}
\end{table}

%% file: new_table_tex/nir-testset-nij.tex
\begin{table}[ht]
\caption{PMI (in hours) estimated on the \textbf{NIR data} by models trained on Warsaw dataset and \textbf{tested on NIJ} dataset.}
\vspace{-0.3cm}
\footnotesize
\label{tab:testset-nij-nir-data}
\begin{tabular}{@{}lcccccc@{}}
\toprule
\multirow{2}{*}{\begin{tabular}[c]{@{}l@{}}Model\\ Backbone\end{tabular}} & \multicolumn{2}{c}{\begin{tabular}[c]{@{}c@{}}Without \\ balancing\end{tabular}} & \multicolumn{2}{c}{\begin{tabular}[c]{@{}c@{}}{\it S3-real}\\ balancing\end{tabular}} & \multicolumn{2}{c}{\begin{tabular}[c]{@{}c@{}}{\it S3-synthetic}\\ balancing\end{tabular}} \\ \cmidrule(l){2-3} \cmidrule(l){4-5} \cmidrule(l){6-7}
 & RMSE & MAE & RMSE & MAE & RMSE & MAE \\ \cmidrule(r){1-1}
VGG19 & 165.37 & 76.67 & 202.05 & 78.96 & 133.06 & 60.11 \\
Inception\_v3 & 170.66 & 80.83 & 206.95 & 79.59 & \textbf{117.31} & \textbf{55.84} \\
DenseNet121 & 171.38 & 77.82 & 206.74 & 78.91 & 133.8 & 56.28 \\
ResNet152 & 163.27 & 76.36 & 207.84 & 79.65 & 130.68 & 59.10 \\ 
ViT & 212.22 & 81.89 & 192.80 & 81.75 & 222.91 & 84.33 \\ 
DS-ResNet152 & 201.06 & 78.59 & 198.27 & 77.30 & 136.49 & 62.27 \\
\bottomrule
\end{tabular}
\end{table}

%% file: new_table_tex/rgb-testset-nij.tex
\begin{table}[ht]
\caption{Same as in Tab. \ref{tab:testset-nij-nir-data} for \textbf{RGB data}.}
\vspace{-0.3cm}
\footnotesize
\label{tab:testset-nij-rgb-data}
\begin{tabular}{@{}lcccccc@{}}
\toprule
\multirow{2}{*}{\begin{tabular}[c]{@{}l@{}}Model\\ Backbone\end{tabular}} & \multicolumn{2}{c}{\begin{tabular}[c]{@{}c@{}}Without \\ balancing\end{tabular}} & \multicolumn{2}{c}{\begin{tabular}[c]{@{}c@{}}{\it S3-real}\\ balancing\end{tabular}} & \multicolumn{2}{c}{\begin{tabular}[c]{@{}c@{}}{\it S3-synthetic}\\ balancing\end{tabular}} \\ \cmidrule(l){2-3} \cmidrule(l){4-5} \cmidrule(l){6-7}
 & RMSE & MAE & RMSE & MAE & RMSE & MAE \\ \cmidrule(r){1-1}
VGG19 & 98.34 & 80.59 & 154.57 & 115.47 & 98.8 & 67.26 \\
Inception\_v3 & 97.34 & 81.1 & 114.26 & 83.6 & \textbf{83.83} & \textbf{54.94} \\
DenseNet121 & 95.24 & 76.32 & 104.97 & 76.36 & 84.92 & 55.27 \\
ResNet152 & 95.31 & 80.11 & 101.38 & 74.37 & 86.67 & 58.91 \\ 
ViT & 98.61 & 72.72 & 97.89 & 77.89 & 103.77 & 76.39 \\ 
DS-ResNet152 & 98.38 & 76.58 & 103.32 & 84.41 & 81.55 & 57.67 \\
\bottomrule
\end{tabular}
\end{table}

%% file: new_table_tex/multi-testset-nij.tex
\begin{table}[ht]
\caption{Same as in Tab. \ref{tab:testset-nij-nir-data} for \textbf{Multispectral data}.}
\vspace{-0.3cm}
\footnotesize
\label{tab:testset-nij-multi-data}
\begin{tabular}{@{}lcccccc@{}}
\toprule
\multirow{2}{*}{\begin{tabular}[c]{@{}l@{}}Model\\ Backbone\end{tabular}} & \multicolumn{2}{c}{\begin{tabular}[c]{@{}c@{}}Without \\ balancing\end{tabular}} & \multicolumn{2}{c}{\begin{tabular}[c]{@{}c@{}}{\it S3-real}\\ balancing\end{tabular}} & \multicolumn{2}{c}{\begin{tabular}[c]{@{}c@{}}{\it S3-synthetic}\\ balancing\end{tabular}} \\ \cmidrule(l){2-3} \cmidrule(l){4-5} \cmidrule(l){6-7}
 & RMSE & MAE & RMSE & MAE & RMSE & MAE \\ \cmidrule(r){1-1}
VGG19 & 152.56 & 85.46 & 166.96 & 80.63 & 108.53 & 61.86 \\
Inception\_v3 & 152.82 & 81.32 & 176.58 & 72.32 & 102.94 & 47.71 \\
DenseNet121 & 136 & 68.55 & 162.44 & 71.29 & 93.57 & 46.58 \\
ResNet152 & 152.63 & 77.66 & 176.76 & 70.91 & \textbf{92.18} & \textbf{45.77} \\ 
ViT & 159.31 & 76.41 & 167.02 & 98.90 & 133.81 & 80.41 \\
DS-ResNet152 & 169.43 & 72.76 & 163.89 & 70.44 & 110.31 & 49.66 \\
\bottomrule
\end{tabular}
\end{table}

%% file: new_figure_tex/war-multi-best-model.tex
\begin{figure}[ht] 
    \centering
    \begin{subfigure}[b]{0.48\linewidth}
        \centering
        \includegraphics[width=\linewidth]{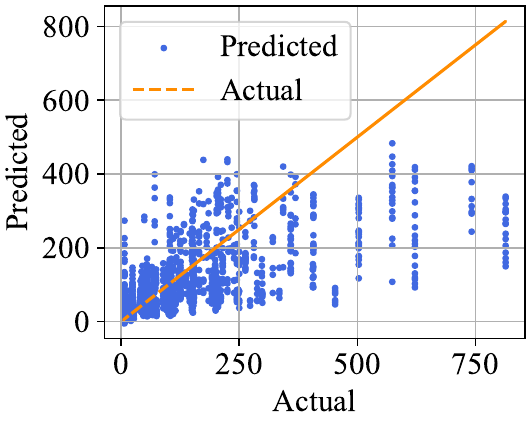}
        \caption{No training data \\balancing}
        \label{warsaw-multi:no-balancing}
    \end{subfigure}%
    \begin{subfigure}[b]{0.48\linewidth}
        \includegraphics[width=\linewidth]{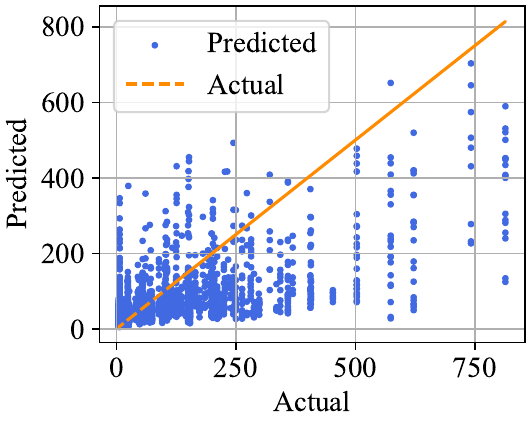}
        \centering
        \caption{Data balancing: up-sampling \\with real data ({\it S3-real}).}
        \label{warsaw-multi:resampling}
    \end{subfigure}%
    \\
    \begin{subfigure}[b]{0.48\linewidth}
        \centering
        \includegraphics[width=\linewidth]{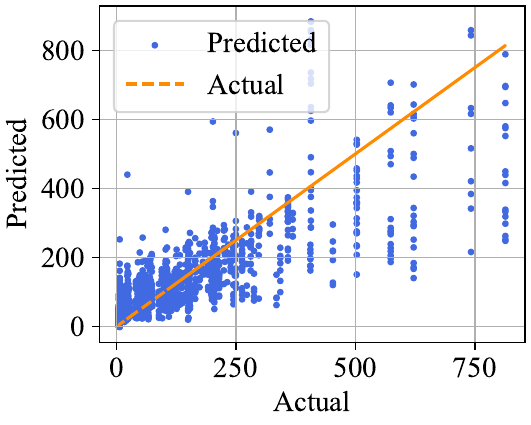}
        \caption{Data balancing: supplementing \\with synthetic data ({\it S3-synthetic}).}
        \label{warsaw-multi:synthetic}
    \end{subfigure}%
    \vspace{-0.2cm}
    \caption{Scatter plots visualizing the predicted PMI values against the actual PMI values for \textbf{Multispectral data} by best performing model (Inception\_v3) trained on the NIJ dataset and \textbf{tested on the Warsaw dataset} and for three approaches to training data balancing.}
    \label{fig:warsaw-multi-best-model-scatter-plot} 
\end{figure} 

%% file: new_figure_tex/nij-multi-best-model.tex
\begin{figure}[ht] 
    \centering
    \begin{subfigure}[b]{0.48\linewidth}
        \centering
        \includegraphics[width=\linewidth]{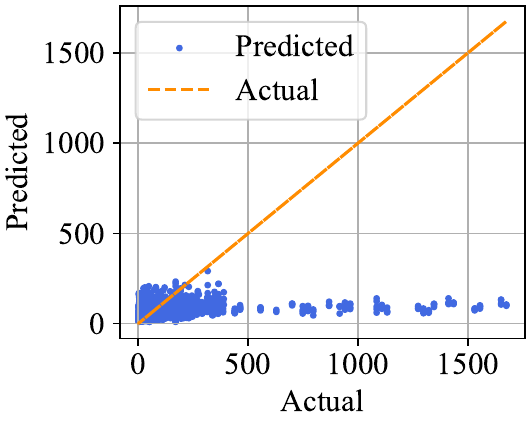}
        \caption{No training data \\balancing}
        \label{nij-multi:no-balancing}
    \end{subfigure}%
    \begin{subfigure}[b]{0.48\linewidth}
        \centering
        \includegraphics[width=\linewidth]{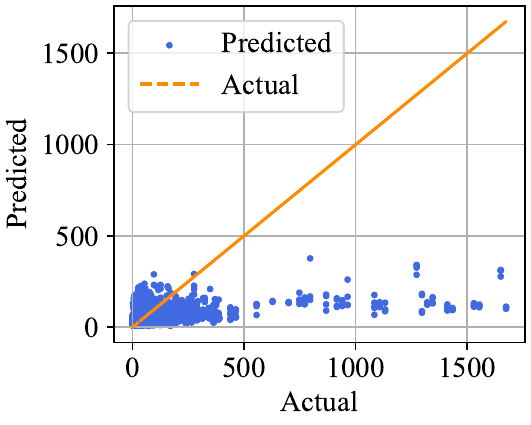}
        \caption{Data balancing: up-sampling \\with real data ({\it S3-real}).}
        \label{nij-multi:resampling}
    \end{subfigure}%
    \\
    \begin{subfigure}[b]{0.48\linewidth}
        \centering
        \includegraphics[width=\linewidth]{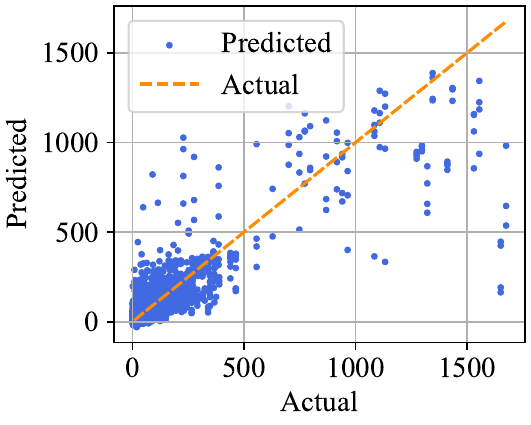}
        \caption{Data balancing: supplementing \\with synthetic data ({\it S3-synthetic}).}
        \label{nij-multi:synthetic}
    \end{subfigure}%
    \vspace{-0.2cm}
    \caption{Same as in Fig. \ref{fig:warsaw-multi-best-model-scatter-plot}, except that results for the best-performing Resnet152 model -- trained on the Warsaw dataset and \textbf{tested on the NIJ dataset} -- are presented.}
    \label{fig:nij-multi-best-model-scatter-plot} 
\end{figure}

%% file: sections/discussion.tex
\paragraph{Commentary to the observed results:} Our study delved, to our knowledge for the first time, into the feasibility of using solely iris image-based data for estimating the PMI, employing forensic iris images representing various spectral bands. Our findings suggest that the PMI estimation, particularly with multispectral data, is feasible and its usefulness depends on the required accuracy.

The first observation made in this study is that dataset-disjoint train-test regime (S3) allows to train models that are almost two orders of magnitude worse than those obtained in sample-disjoint setup (S1), and approx. four times worse than those trained according to the subject-disjoint train-test protocol (S2). This shows how difficult the automatic estimation of the PMI can be if one poses solely the iris image and the evaluation corresponds to a real application (data collected by disjoint labs).

The second observation is that a narrower range of PMI for samples included into Warsaw data (cf. Fig. \ref{fig:dataset_box_plot}) results in significantly worse models, what is illustrated in scatter plots \ref{fig:warsaw-multi-best-model-scatter-plot} and \ref{fig:nij-multi-best-model-scatter-plot}: models trained solely on the Warsaw samples (see Fig. \ref{fig:nij-multi-best-model-scatter-plot}) are unable to predict correct PMI on the NIJ dataset, while models trained solely on the NIJ samples (see Fig. \ref{fig:warsaw-multi-best-model-scatter-plot}) present much better capabilities to predict the PMI on the Warsaw dataset. \textcolor{black}{This is likely because a larger variance of PMI in the training data allows for better generalization.} These observations, made for multispectral data-based training, are true also for models trained with NIR and RGB data, and the corresponding plots are included in the supplementary materials (Figures \ref{fig:warsaw-nir-best-model-scatter-plot}, \ref{fig:nij-nir-best-model-scatter-plot}, \ref{fig:warsaw-rgb-best-model-scatter-plot} and \ref{fig:nij-rgb-best-model-scatter-plot}).

The third, and very positive observation is that supplementing the always-limited and not-representative (in terms of PMI) real data with synthetically-generated forensic iris images (with a generative model conditioned by the PMI) allows to reduce the estimation error (MAE) from approx. 3 days (77.66 hours) to approx. 2 days (45.77 hours), see Tab. \ref{tab:testset-nij-multi-data}. With constant improvement of generative models and their abilities to correctly model intricate biological processes as the iris decomposition, this bodes well for future improvements of this approach.

\vskip1mm\noindent
{\bf Applications:} 
This is the first-of-it-kind attempt to use solely the iris image to estimate the PMI. One direct and obvious application is to serve as an element of the forensic toolkit, in which the PMI estimation may integrate several approaches, including those not related to the use of biometric data. The second, biometrics-related application, is to use the estimated PMI in conditioning post-mortem iris recognition methods, which were proposed in the past \cite{Trokielewicz_IVC_2020,Trokielewicz_WACV_2020,Boyd_Access_2020} but without considering PMI as a factor guiding the classifiers to using different features depending on the decomposition state. 

\vskip1mm\noindent
{\bf Limitations:}
First, relying solely on iris images may overlook crucial contextual factors like ambient temperature fluctuations or post-mortem changes in other tissues, potentially useful in PMI estimation. Addressing this by incorporating additional contextual information could enhance the robustness of our models. \textcolor{black}{Second, other biological factors, such as gender, age, death reason, and dietary habits, may introduce potential bias.} Third, generalization from our findings may be hampered by the dataset limited size and diversity despite rigorous cross-validation. Recent growth of generative models, including those designed specifically for post-mortem iris images \cite{Bhuiyan_WACVW_2024}, may partially and soon solve this limitation.

%% file: sections/supplementary.tex

\onecolumn{%
\centering
{\large \textbf{Forensic Iris Image-Based Post-Mortem Interval Estimation}\\\vskip1mm 
Supplementary Materials}\vskip3mm

\input{figure_tex/10f_nir_box_plot}
\input{figure_tex/10f_rgb_box_plot}

\input{figure_tex/subject_disjoint_10f_nir_box_plot}

\input{figure_tex/subject_disjoint_10f_rgb_box_plot}
}

\input{figure_tex/10f_nir_model_plot}
\input{figure_tex/10f_rgb_model_plot}

\input{figure_tex/subject_disjoint_10f_nir_model_plot}
\input{figure_tex/subject_disjoint_10f_rgb_model_plot}

\input{new_figure_tex/war-nir-best-model}
\input{new_figure_tex/nij-nir-best-model}

\input{new_figure_tex/war-rgb-best-model}
\input{new_figure_tex/nij-rgb-best-model}

%% file: figure_tex/10f_nir_box_plot.tex
\begin{figure}[htpb] 
    \centering
    \begin{subfigure}[b]{0.3\linewidth}
        \centering
        \includegraphics[width=\linewidth]{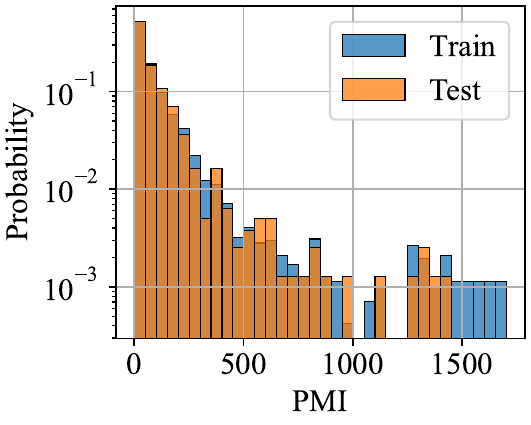}
        \caption{Least performance fold}
        \label{10f-nir:least-fold}
    \end{subfigure}%
    \begin{subfigure}[b]{0.3\linewidth}
        \centering
        \includegraphics[width=\linewidth]{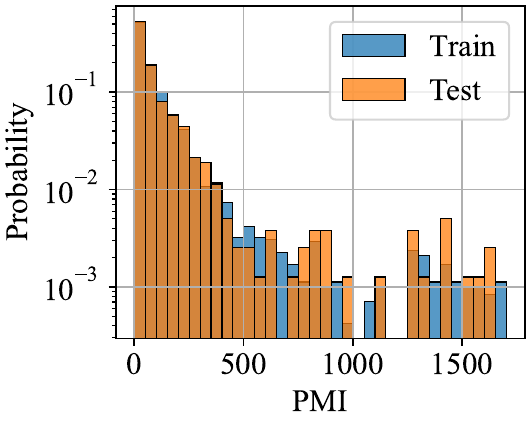}
        \caption{Best performance fold}
        \label{10f-nir:best-fold}
    \end{subfigure}%
    \caption{Comparison of the PMI distributions between training and test datasets for the least (a) and best (b) performing folds observed in the \textbf{sample-disjoint} 10-fold cross-validation for the \textbf{NIR data}..}
    \label{fig:nir_10f_box_plot} 
\end{figure} 

%% file: figure_tex/10f_rgb_box_plot.tex
\begin{figure}[!htpb] 
    \centering
    \begin{subfigure}[b]{0.3\linewidth}
        \centering
        \includegraphics[width=\linewidth]{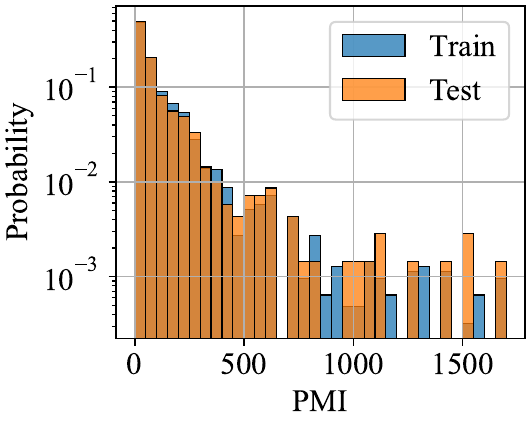}
        \caption{Least performance fold}
        \label{10f-rgb:least-fold}
    \end{subfigure}%
    \begin{subfigure}[b]{0.3\linewidth}
        \centering
        \includegraphics[width=\linewidth]{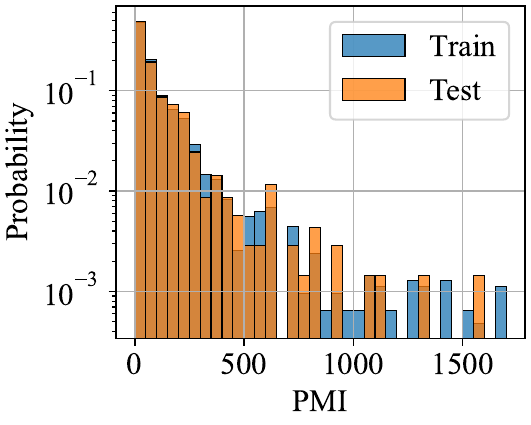}
        \caption{Best performance fold}
        \label{10f-rgb:best-fold}
    \end{subfigure}%
    \caption{Same as in Fig. \ref{fig:nir_10f_box_plot} but for \textbf{RGB data}.}
    \label{fig:rgb_10f_box_plot} 
\end{figure}

%% file: figure_tex/subject_disjoint_10f_nir_box_plot.tex
\begin{figure}[!htpb] 
    \centering
    \begin{subfigure}[b]{0.3\linewidth}
        \centering
        \includegraphics[width=\linewidth]{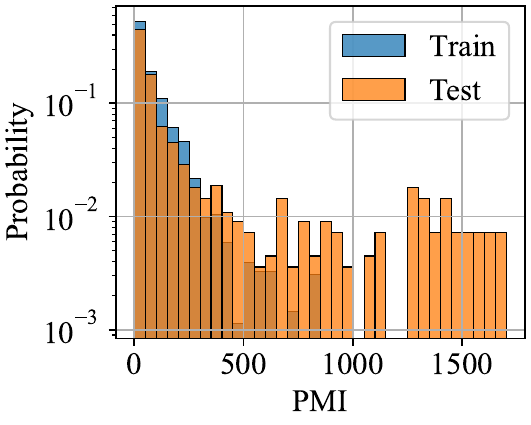}
        \caption{Least performance fold}
        \label{subject-nir:least-fold}
    \end{subfigure}%
    \begin{subfigure}[b]{0.3\linewidth}
        \centering
        \includegraphics[width=\linewidth]{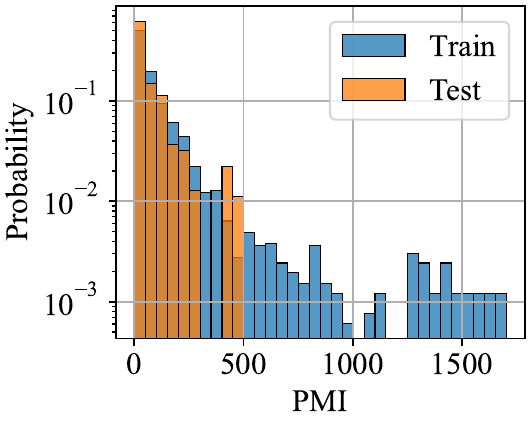}
        \caption{Best performance fold}
        \label{subject-nir:best-fold}
    \end{subfigure}%
    \caption{Comparison of the PMI distributions between training and test datasets for the least (a) and best (b) performing folds observed in the \textbf{subject-disjoint} 10-fold cross-validation for the \textbf{NIR data}.}
    \label{fig:nir_subject_box_plot} 
\end{figure}

%% file: figure_tex/subject_disjoint_10f_rgb_box_plot.tex
\begin{figure}[!htpb] 
    \centering
    \begin{subfigure}[b]{0.3\linewidth}
        \centering
        \includegraphics[width=\linewidth]{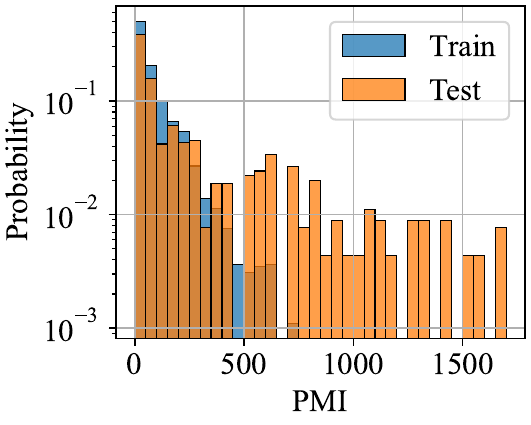}
        \caption{Least performance fold}
        \label{subject-rgb:least-fold}
    \end{subfigure}%
    \begin{subfigure}[b]{0.3\linewidth}
        \centering
        \includegraphics[width=\linewidth]{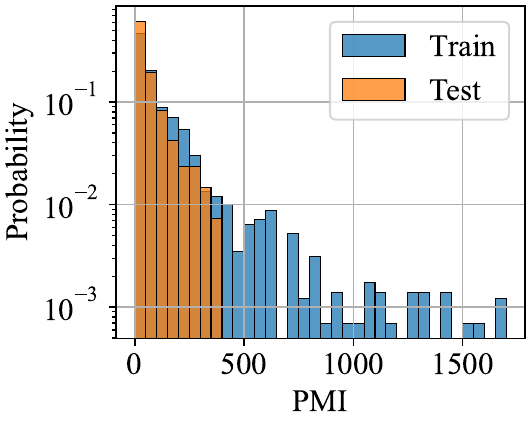}
        \caption{Best performance fold}
        \label{subject-rgb:best-fold}
    \end{subfigure}%
    \caption{Same as in Fig. \ref{fig:nir_subject_box_plot} but for \textbf{RGB data}.}
    \label{fig:rgb_subject_box_plot} 
\end{figure} 

%% file: figure_tex/10f_nir_model_plot.tex
\begin{figure*}[ht] 
    \centering
    \begin{subfigure}[b]{0.25\linewidth}
        \centering
        \includegraphics[width=\linewidth]{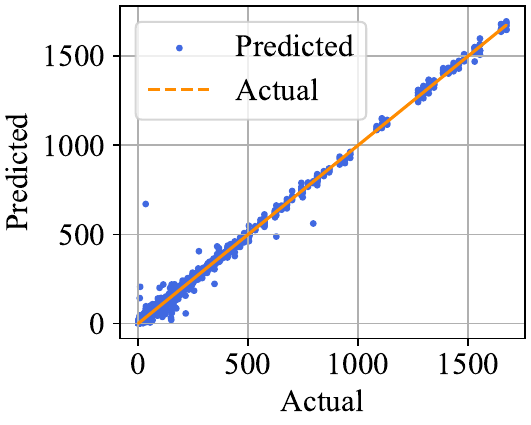}
        \caption{VGG19}
        \label{nir:vgg19}
    \end{subfigure}%
    \begin{subfigure}[b]{0.25\linewidth}
        \centering
        \includegraphics[width=\linewidth]{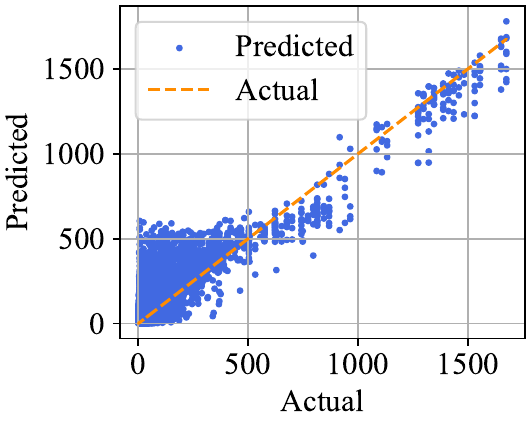}
        \caption{Inception\_v3}
        \label{nir:inception}
    \end{subfigure}%
    \begin{subfigure}[b]{0.25\linewidth}
        \centering
        \includegraphics[width=\linewidth]{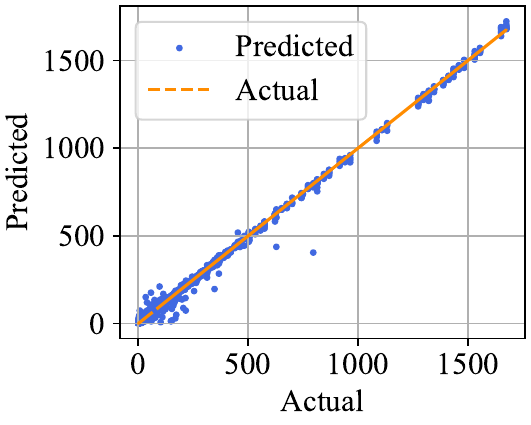}
        \caption{DenseNet121}
        \label{nir:densenet121}
    \end{subfigure}%
    \begin{subfigure}[b]{0.25\linewidth}
        \centering
        \includegraphics[width=\linewidth]{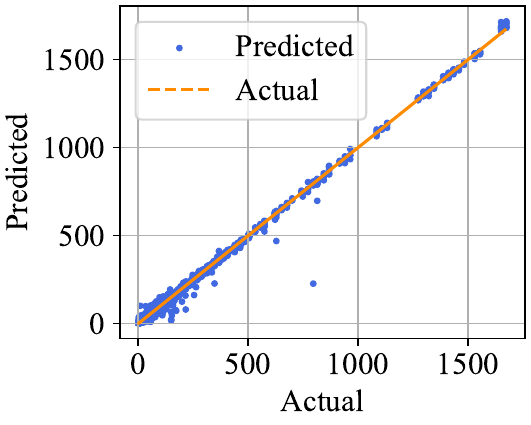}
        \caption{ResNet152}
        \label{nir:resnet152}
    \end{subfigure}
    \caption{Scatter plots visualizing the predicted PMI values against the actual PMI values for \textbf{NIR data} combined for all \textbf{sample-disjoint} 10-fold cross-validations.}
    \label{fig:nir_model_plot} 
\end{figure*} 

%% file: figure_tex/10f_rgb_model_plot.tex
\begin{figure*}[ht] 
    \centering
    \begin{subfigure}[b]{0.25\linewidth}
        \centering
        \includegraphics[width=\linewidth]{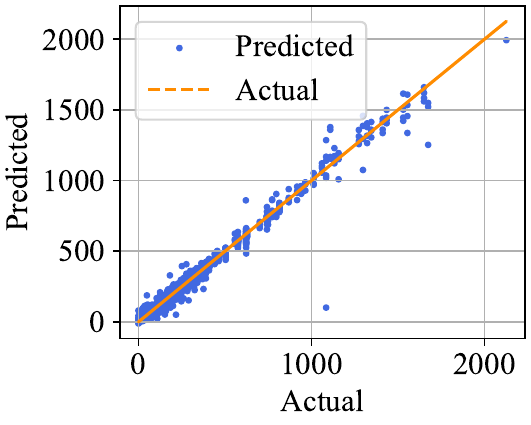}
        \caption{VGG19}
        \label{RGB:vgg19}
    \end{subfigure}%
    \begin{subfigure}[b]{0.25\linewidth}
        \centering
        \includegraphics[width=\linewidth]{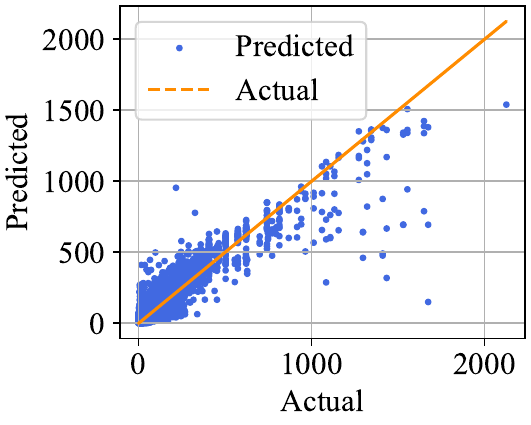}
        \caption{Inception\_v3}
        \label{RGB:inception}
    \end{subfigure}%
    \begin{subfigure}[b]{0.25\linewidth}
        \centering
        \includegraphics[width=\linewidth]{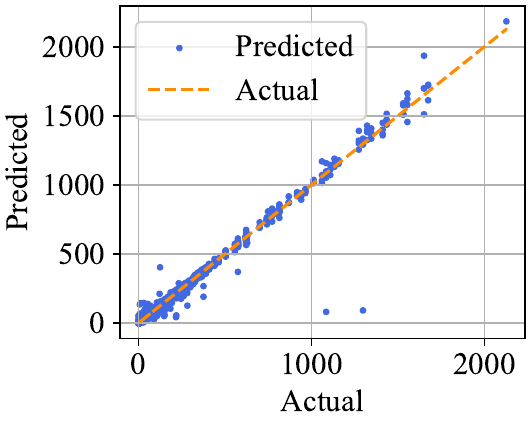}
        \caption{DenseNet121}
        \label{RGB:densenet121}
    \end{subfigure}%
    \begin{subfigure}[b]{0.25\linewidth}
        \centering
        \includegraphics[width=\linewidth]{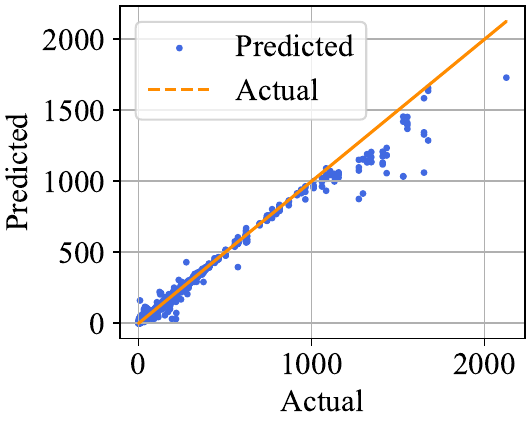}
        \caption{ResNet152}
        \label{RGB:resnet152}
    \end{subfigure}
    \caption{Scatter plots visualizing the predicted PMI values against the actual PMI values for \textbf{RGB data} combined for all \textbf{sample-disjoint} 10-fold cross-validations.}
    \label{fig:RGB_model_plot} 
\end{figure*}

%% file: figure_tex/subject_disjoint_10f_nir_model_plot.tex
\begin{figure*}[ht] 
    \centering
    \begin{subfigure}[b]{0.25\linewidth}
        \centering
        \includegraphics[width=\linewidth]{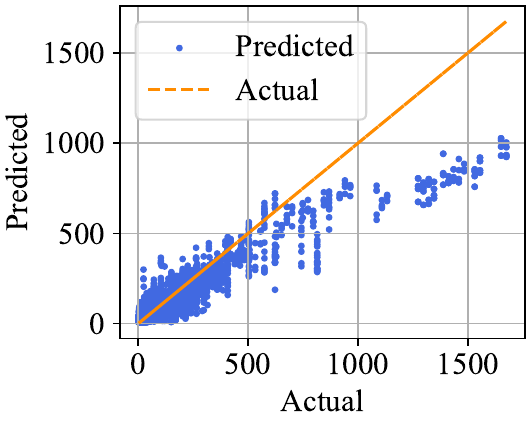}
        \caption{VGG19}
        \label{subject-nir:vgg19}
    \end{subfigure}%
    \begin{subfigure}[b]{0.25\linewidth}
        \centering
        \includegraphics[width=\linewidth]{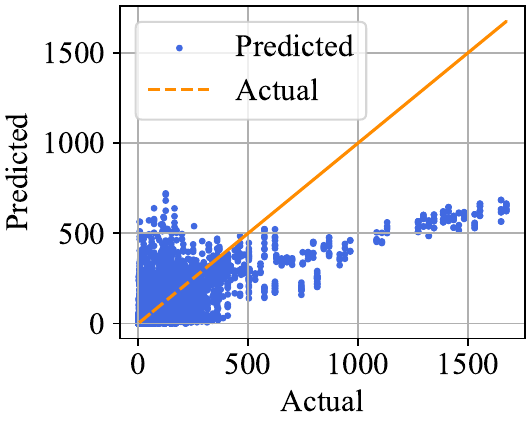}
        \caption{Inception\_v3}
        \label{subject-nir:inception}
    \end{subfigure}%
    \begin{subfigure}[b]{0.25\linewidth}
        \centering
        \includegraphics[width=\linewidth]{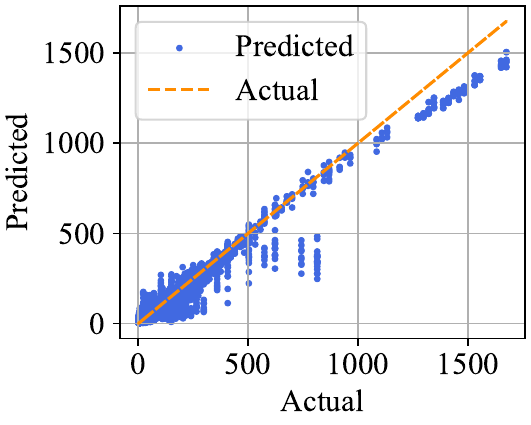}
        \caption{DenseNet121}
        \label{subject-nir:densenet121}
    \end{subfigure}%
    \begin{subfigure}[b]{0.25\linewidth}
        \centering
        \includegraphics[width=\linewidth]{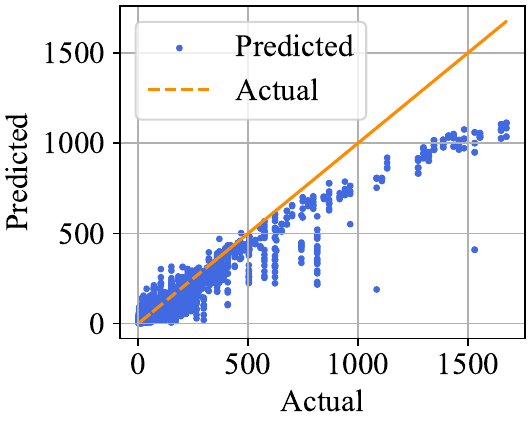}
        \caption{ResNet152}
        \label{subject-nir:resnet152}
    \end{subfigure}
    \caption{Scatter plots visualizing the predicted PMI values against the actual PMI values for \textbf{NIR data} combined for all \textbf{subject-disjoint} 10-fold cross-validations.}
    \label{fig:subject-disjoint-nir} 
\end{figure*} 

%% file: figure_tex/subject_disjoint_10f_rgb_model_plot.tex
\begin{figure*}[ht] 
    \centering
    \begin{subfigure}[b]{0.25\linewidth}
        \centering
        \includegraphics[width=\linewidth]{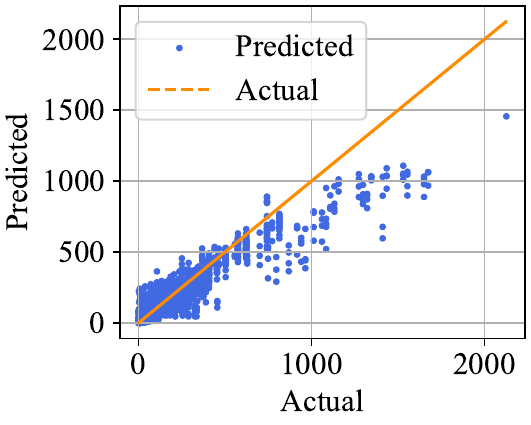}
        \caption{VGG19}
        \label{subject-rgb:vgg19}
    \end{subfigure}%
    \begin{subfigure}[b]{0.25\linewidth}
        \centering
        \includegraphics[width=\linewidth]{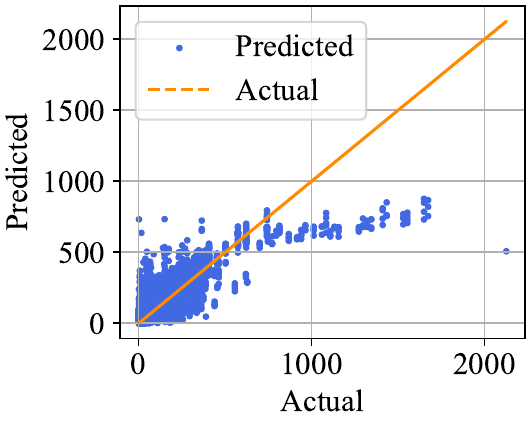}
        \caption{Inception\_v3}
        \label{subject-rgb:inception}
    \end{subfigure}%
    \begin{subfigure}[b]{0.25\linewidth}
        \centering
        \includegraphics[width=\linewidth]{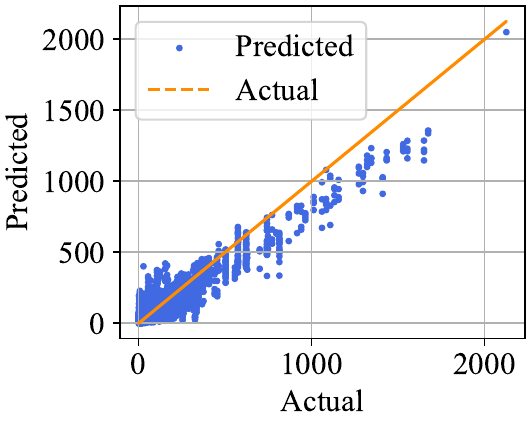}
        \caption{DenseNet121}
        \label{subject-rgb:densenet121}
    \end{subfigure}%
    \begin{subfigure}[b]{0.25\linewidth}
        \centering
        \includegraphics[width=\linewidth]{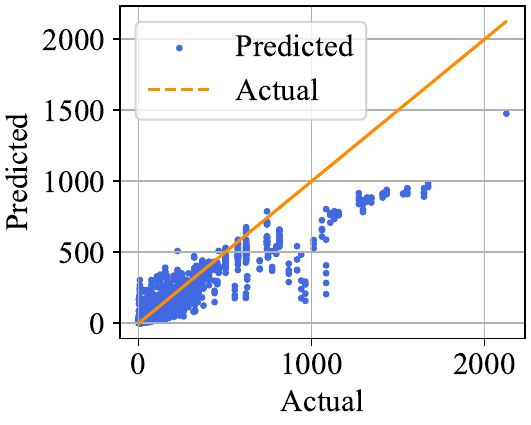}
        \caption{ResNet152}
        \label{subject-rgb:resnet152}
    \end{subfigure}
    \caption{Scatter plots visualizing the predicted PMI values against the actual PMI values for \textbf{RGB data} combined for all \textbf{subject-disjoint} 10-fold cross-validations.}
    \label{fig:subject-disjoint-rgb} 
\end{figure*}

%% file: new_figure_tex/war-nir-best-model.tex
\begin{figure*}[ht] 
    \centering
    \begin{subfigure}[b]{0.25\linewidth}
        \centering
        \includegraphics[width=\linewidth]{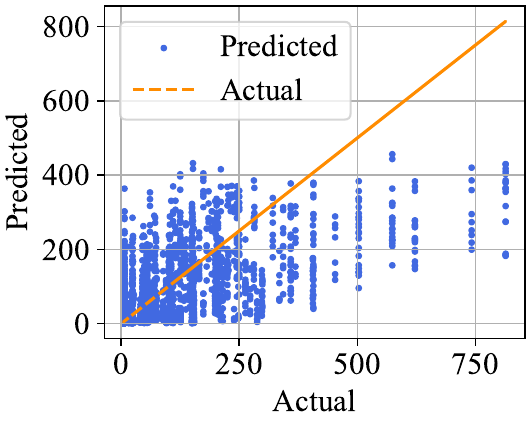}
        \caption{No training data \\balancing}
        \label{warsaw-nir:no-balancing}
    \end{subfigure}%
    \begin{subfigure}[b]{0.25\linewidth}
        \centering
        \includegraphics[width=\linewidth]{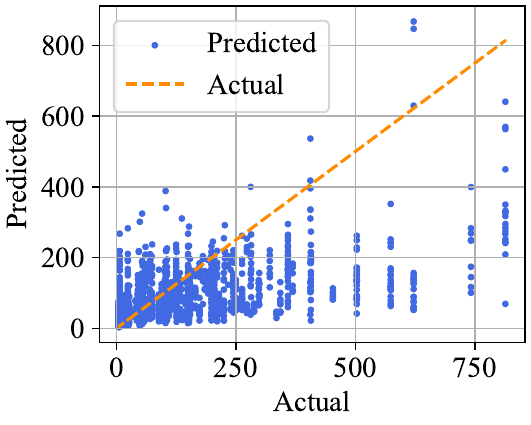}
        \caption{Data balancing: up-sampling \\with real data ({\it S3-real}).}
        \label{warsaw-nir:resampling}
    \end{subfigure}%
    \begin{subfigure}[b]{0.25\linewidth}
        \centering
        \includegraphics[width=\linewidth]{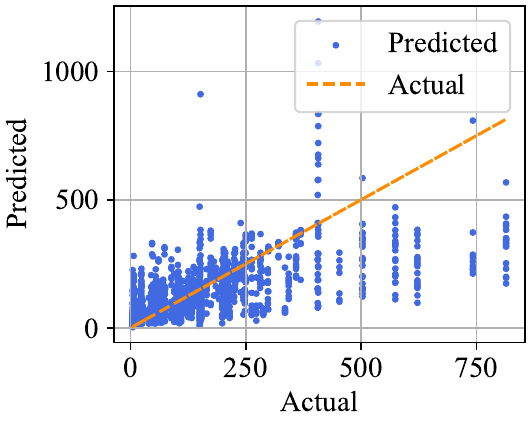}
        \caption{Data balancing: supplementing \\with synthetic data ({\it S3-synthetic}).}
        \label{warsaw-nir:synthetic}
    \end{subfigure}%
    \caption{Scatter plots visualizing the predicted PMI values against the actual PMI values for \textbf{NIR data} by best performing model (Inception\_v3) trained on the NIJ dataset and \textbf{tested on the Warsaw dataset}.}
    \label{fig:warsaw-nir-best-model-scatter-plot} 
\end{figure*} 

%% file: new_figure_tex/nij-nir-best-model.tex
\begin{figure*}[ht] 
    \centering
    \begin{subfigure}[b]{0.25\linewidth}
        \centering
        \includegraphics[width=\linewidth]{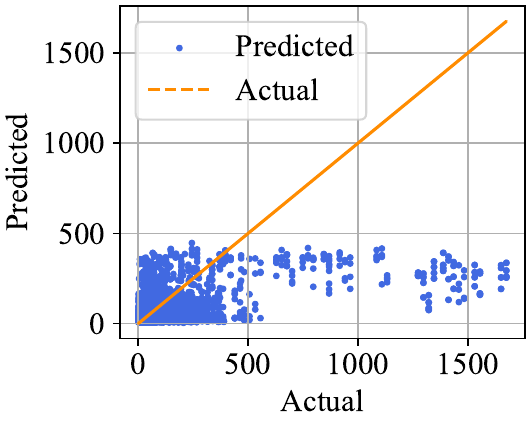}
        \caption{No training data \\balancing}
        \label{nij-nir:no-balancing}
    \end{subfigure}%
    \begin{subfigure}[b]{0.25\linewidth}
        \centering
        \includegraphics[width=\linewidth]{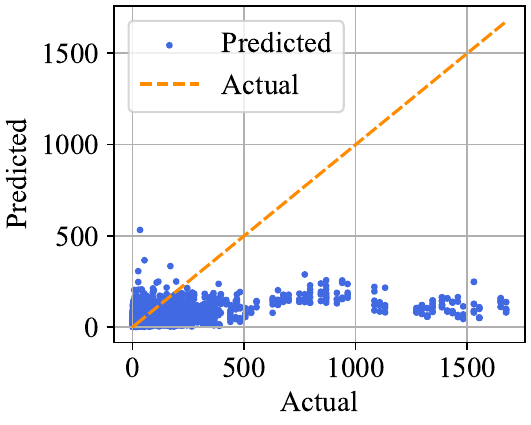}
        \caption{Data balancing: up-sampling \\with real data ({\it S3-real}).}
        \label{nij-nir:resampling}
    \end{subfigure}%
    \begin{subfigure}[b]{0.25\linewidth}
        \centering
        \includegraphics[width=\linewidth]{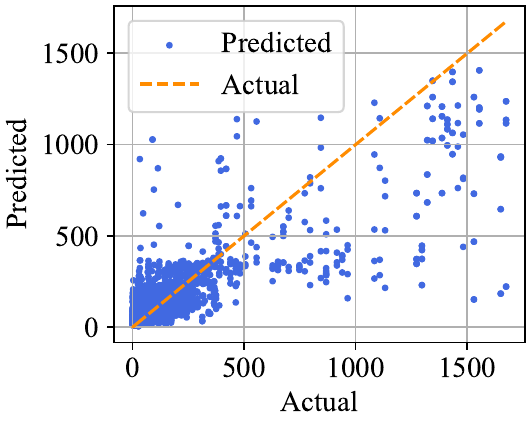}
        \caption{Data balancing: supplementing \\with synthetic data ({\it S3-synthetic}).}
        \label{nij-nir:synthetic}
    \end{subfigure}%
    \caption{Scatter plots visualizing the predicted PMI values against the actual PMI values for \textbf{NIR data} by best performing model (Inception\_v3) trained on the Warsaw dataset and \textbf{tested on the NIJ dataset}.}
    \label{fig:nij-nir-best-model-scatter-plot} 
\end{figure*}

%% file: new_figure_tex/war-rgb-best-model.tex
\begin{figure*}[ht] 
    \centering
    \begin{subfigure}[b]{0.25\linewidth}
        \centering
        \includegraphics[width=\linewidth]{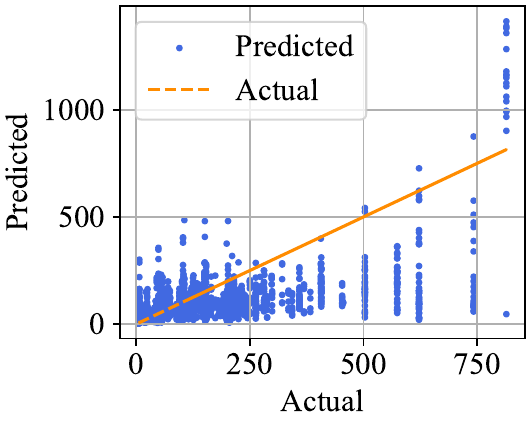}
        \caption{No training data \\balancing}
        \label{warsaw-rgb:no-balancing}
    \end{subfigure}%
    \begin{subfigure}[b]{0.25\linewidth}
        \centering
        \includegraphics[width=\linewidth]{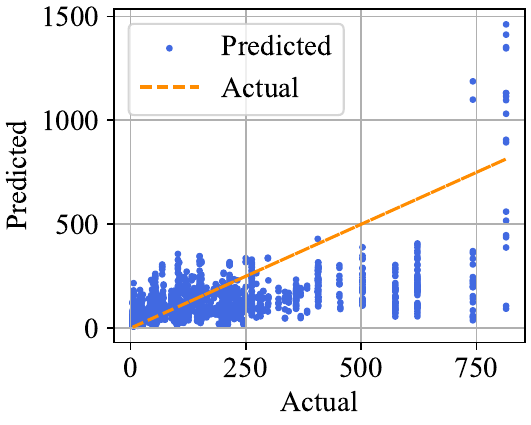}
        \caption{Data balancing: up-sampling \\with real data ({\it S3-real}).}
        \label{warsaw-rgb:resampling}
    \end{subfigure}%
    \begin{subfigure}[b]{0.25\linewidth}
        \centering
        \includegraphics[width=\linewidth]{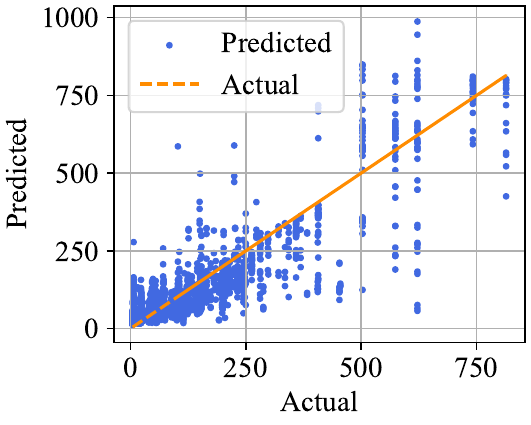}
        \caption{Data balancing: supplementing \\with synthetic data ({\it S3-synthetic}).}
        \label{warsaw-rgb:synthetic}
    \end{subfigure}%
    \caption{Scatter plots visualizing the predicted PMI values against the actual PMI values for \textbf{RGB data} by best performing model (Densenet121) trained on the NIJ dataset and \textbf{tested on the Warsaw dataset}.}
    \label{fig:warsaw-rgb-best-model-scatter-plot} 
\end{figure*} 

%% file: new_figure_tex/nij-rgb-best-model.tex
\begin{figure*}[ht] 
    \centering
    \begin{subfigure}[b]{0.25\linewidth}
        \centering
        \includegraphics[width=\linewidth]{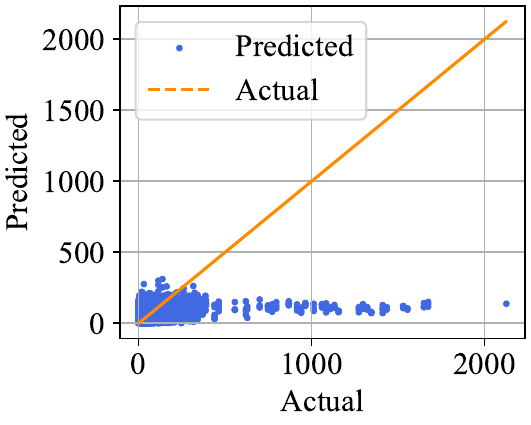}
        \caption{No training data \\balancing}
        \label{nij-rgb:no-balancing}
    \end{subfigure}%
    \begin{subfigure}[b]{0.25\linewidth}
        \centering
        \includegraphics[width=\linewidth]{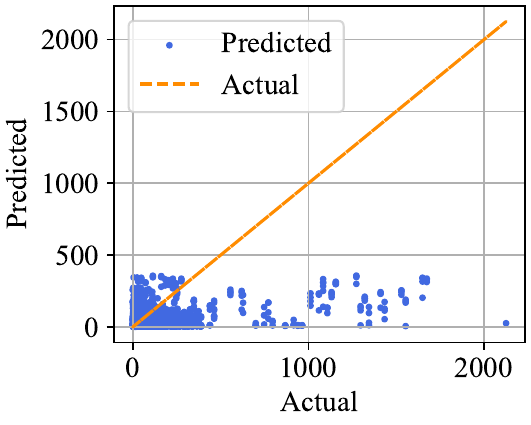}
        \caption{Data balancing: up-sampling \\with real data ({\it S3-real}).}
        \label{nij-rgb:resampling}
    \end{subfigure}%
    \begin{subfigure}[b]{0.25\linewidth}
        \centering
        \includegraphics[width=\linewidth]{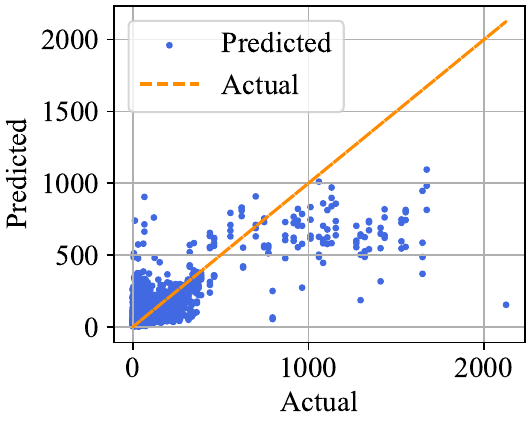}
        \caption{Data balancing: supplementing \\with synthetic data ({\it S3-synthetic}).}
        \label{nij-rgb:synthetic}
    \end{subfigure}%
    \caption{Scatter plots visualizing the predicted PMI values against the actual PMI values for \textbf{RGB data} by best performing model (Inception\_v3) trained on the Warsaw dataset and \textbf{tested on the NIJ dataset}.}
    \label{fig:nij-rgb-best-model-scatter-plot} 
\end{figure*}

%% file: main.bbl
\begin{thebibliography}{10}\itemsep=-1pt

\bibitem{FBI_NGI_webpage}
{FBI Next Generation Identification (NGI)}.
\newblock \url{https://le.fbi.gov/science-and-lab/biometrics-and-fingerprints/biometrics/next-generation-identification-ngi}.
\newblock Accessed: March 24, 2024.

\bibitem{Boyd_Access_2020}
Aidan Boyd, Shivangi Yadav, Thomas Swearingen, Andrey Kuehlkamp, Mateusz Trokielewicz, Eric Benjamin, Piotr Maciejewicz, Dennis Chute, Arun Ross, Patrick Flynn, Kevin Bowyer, and Adam Czajka.
\newblock {Post-Mortem Iris Recognition -- A Survey and Assessment of the State of the Art}.
\newblock {\em IEEE Access}, 8:136570--136593, 2020.

\bibitem{daugman2001bbc}
John Daugman.
\newblock Bbc news: The eyes have it.
\newblock \url{http://news.bbc.co.uk/2/hi/science/nature/1477655.stm}, 2001.
\newblock [Accessed: 2021-08-14].

\bibitem{szczepanski2014pupil}
Adam Szczepa{\'n}ski, Krzysztof Misztal, and Khalid Saeed.
\newblock Pupil and iris detection algorithm for near-infrared capture devices.
\newblock In {\em Computer Information Systems and Industrial Management: 13th IFIP TC8 International Conference, CISIM 2014, Ho Chi Minh City, Vietnam, November 5-7, 2014. Proceedings 14}, pages 141--150. Springer, 2014.

\bibitem{IrisGuard2016}
IrisGuard.
\newblock Eyebank solution.
\newblock \url{http://www.irisguard.com/eyebank/downloads/EyeBankPer\%20Page.pdf}, 2016.
\newblock [accessed: Jan 2016].

\bibitem{IriTech2015}
IriTech.
\newblock Biometric access control can iris biometric enhance better security?
\newblock \url{http://www.iritech.com/blog/iris-biometric-access-control}, August 20, 2015.
\newblock [accessed: January 16, 2016].

\bibitem{bolme2016impact}
David~S Bolme, Ryan~A Tokola, Chris~B Boehnen, Tiffany~B Saul, Kelly~A Sauerwein, and Dawnie~Wolfe Steadman.
\newblock Impact of environmental factors on biometric matching during human decomposition.
\newblock In {\em 2016 IEEE 8th International Conference on Biometrics Theory, Applications and Systems (BTAS)}, pages 1--8. IEEE, 2016.

\bibitem{sauerwein2017effect}
Kelly Sauerwein, Tiffany~B Saul, Dawnie~Wolfe Steadman, and Chris~B Boehnen.
\newblock The effect of decomposition on the efficacy of biometrics for positive identification.
\newblock {\em Journal of Forensic Sciences}, 62(6):1599--1602, 2017.

\bibitem{trokielewicz2018iris}
Mateusz Trokielewicz, Adam Czajka, and Piotr Maciejewicz.
\newblock Iris recognition after death.
\newblock {\em IEEE Transactions on Information Forensics and Security}, 14(6):1501--1514, 2018.

\bibitem{kaliszan2009estimation}
Micha{\l} Kaliszan, Roman Hauser, and Gerhard Kernbach-Wighton.
\newblock Estimation of the time of death based on the assessment of post mortem processes with emphasis on body cooling.
\newblock {\em Legal Medicine}, 11(3):111--117, 2009.

\bibitem{rodrigo2015time}
Marianito~R Rodrigo.
\newblock Time of death estimation from temperature readings only: A laplace transform approach.
\newblock {\em Applied Mathematics Letters}, 39:47--52, 2015.

\bibitem{igari2016rectal}
Yui Igari, Yoshiyuki Hosokai, and Masato Funayama.
\newblock Rectal temperature-based death time estimation in infants.
\newblock {\em Legal Medicine}, 19:35--42, 2016.

\bibitem{rodrigo2016nonlinear}
Marianito~R Rodrigo.
\newblock A nonlinear least squares approach to time of death estimation via body cooling.
\newblock {\em Journal of Forensic Sciences}, 61(1):230--233, 2016.

\bibitem{varetto2005long}
Lorenzo Varetto and Ombretta Curto.
\newblock Long persistence of rigor mortis at constant low temperature.
\newblock {\em Forensic Science International}, 147(1):31--34, 2005.

\bibitem{ozawa2013effect}
Masayoshi Ozawa, Kimiharu Iwadate, Sari Matsumoto, Kumiko Asakura, Eriko Ochiai, and Kyoko Maebashi.
\newblock The effect of temperature on the mechanical aspects of rigor mortis in a liquid paraffin model.
\newblock {\em Legal Medicine}, 15(6):293--297, 2013.

\bibitem{nishida2015blood}
Atsushi Nishida, Hironao Funaki, Masaki Kobayashi, Yuka Tanaka, Yoshihisa Akasaka, Toshikazu Kubo, and Hiroshi Ikegaya.
\newblock Blood creatinine level in postmortem cases.
\newblock {\em Science \& Justice}, 55(3):195--199, 2015.

\bibitem{martins2015necromechanics}
Pedro~ALS Martins, Francisca Ferreira, Renato Natal~Jorge, Marco Parente, and Agostinho Santos.
\newblock Necromechanics: death-induced changes in the mechanical properties of human tissues.
\newblock {\em Proceedings of the Institution of Mechanical Engineers, Part H: Journal of Engineering in Medicine}, 229(5):343--349, 2015.

\bibitem{vass2002decomposition}
Arpad~A Vass, Stacy-Ann Barshick, Gary Sega, John Caton, James~T Skeen, Jennifer~C Love, and Jennifer~A Synstelien.
\newblock Decomposition chemistry of human remains: a new methodology for determining the postmortem interval.
\newblock {\em Journal of Forensic Sciences}, 47(3):542--553, 2002.

\bibitem{ferreira2013can}
M~Teresa Ferreira and Eug{\'e}nia Cunha.
\newblock Can we infer post mortem interval on the basis of decomposition rate? a case from a portuguese cemetery.
\newblock {\em Forensic Science International}, 226(1-3):298--e1, 2013.

\bibitem{cockle2015human}
Diane~L Cockle and Lynne~S Bell.
\newblock Human decomposition and the reliability of a ‘universal’model for post mortem interval estimations.
\newblock {\em Forensic Science International}, 253:136--e1, 2015.

\bibitem{suckling2016longitudinal}
Joanna~K Suckling, M~Katherine Spradley, and Kanya Godde.
\newblock A longitudinal study on human outdoor decomposition in central texas.
\newblock {\em Journal of Forensic Sciences}, 61(1):19--25, 2016.

\bibitem{chandrakanth2013postmortem}
HV Chandrakanth, Tanuj Kanchan, BM Balaraj, HS Virupaksha, and TN Chandrashekar.
\newblock Postmortem vitreous chemistry--an evaluation of sodium, potassium and chloride levels in estimation of time since death (during the first 36 h after death).
\newblock {\em Journal of Forensic and Legal Medicine}, 20(4):211--216, 2013.

\bibitem{cordeiro2015application}
Cristina Cordeiro, Rafael Seoane, Ana Camba, Elena Lendoiro, Mar{\'\i}a~S Rodr{\'\i}guez-Calvo, Duarte~N Vieira, and Jos{\'e}~I Mu{\~n}oz-Bar{\'u}s.
\newblock The application of flow cytometry as a rapid and sensitive screening method to detect contamination of vitreous humor samples and avoid miscalculation of the postmortem interval.
\newblock {\em Journal of Forensic Sciences}, 60(5):1346--1349, 2015.

\bibitem{zilg2015new}
B Zilg, Samuel Bernard, K Alkass, S{\"o}ren Berg, and H Druid.
\newblock A new model for the estimation of time of death from vitreous potassium levels corrected for age and temperature.
\newblock {\em Forensic science international}, 254:158--166, 2015.

\bibitem{parmar2015estimation}
Ankita~K Parmar and Shobhana~K Menon.
\newblock Estimation of postmortem interval through albumin in csf by simple dye binding method.
\newblock {\em Science \& Justice}, 55(6):388--393, 2015.

\bibitem{rognum2016estimation}
TO Rognum, S Holmen, MA Musse, PS Dahlberg, A Stray-Pedersen, OD Saugstad, and SH Opdal.
\newblock Estimation of time since death by vitreous humor hypoxanthine, potassium, and ambient temperature.
\newblock {\em Forensic Science International}, 262:160--165, 2016.

\bibitem{tuccia2016combined}
Fabiola Tuccia, Giorgia Giordani, and Stefano Vanin.
\newblock A combined protocol for identification of maggots of forensic interest.
\newblock {\em Science \& Justice}, 56(4):264--268, 2016.

\bibitem{iancu2016dynamics}
Lavinia Iancu, Tiberiu Sahlean, and Cristina Purcarea.
\newblock Dynamics of necrophagous insect and tissue bacteria for postmortem interval estimation during the warm season in romania.
\newblock {\em Journal of Medical Entomology}, 53(1):54--66, 2016.

\bibitem{canturk2016experimental}
{\.I}smail Cant{\"u}rk, Fethullah Karabiber, Safa {\c{C}}elik, M~Feyzi {\c{S}}ahin, Fatih Ya{\u{g}}mur, and Sad{\i}k Kara.
\newblock An experimental evaluation of electrical skin conductivity changes in postmortem interval and its assessment for time of death estimation.
\newblock {\em Computers in Biology and Medicine}, 69:92--96, 2016.

\bibitem{kumar2012determination}
Binay Kumar, Vinita Kumari, Tulsi Mahto, Ashok Sharma, and Aman Kumar.
\newblock Determination of time elapsed since death from the status of transparency of cornea in ranchi in different weathers.
\newblock {\em Journal of Indian Academy of Forensic Medicine}, 34(4):336--338, 2012.

\bibitem{zhou2010image}
Lan Zhou, Yan Liu, Liang Liu, Luo Zhuo, Man Liang, Fan Yang, Liang Ren, and Shaohua Zhu.
\newblock Image analysis on corneal opacity: a novel method to estimate postmortem interval in rabbits.
\newblock {\em Journal of Huazhong University of Science and Technology [Medical Sciences]}, 30:235--239, 2010.

\bibitem{liu2008image}
Fang Liu, Shaohua Zhu, Yuxiao Fu, Fan Fan, Tianjiang Wang, and Songfeng Lu.
\newblock Image analysis of the relationship between changes of cornea and postmortem interval.
\newblock In {\em PRICAI 2008: Trends in Artificial Intelligence: 10th Pacific Rim International Conference on Artificial Intelligence, Hanoi, Vietnam, December 15-19, 2008. Proceedings 10}, pages 998--1003. Springer, 2008.

\bibitem{kawashima2015estimating}
Wataru Kawashima, Katsuhiko Hatake, Risa Kudo, Mari Nakanishi, Shigehiro Tamaki, Shogo Kasuda, and A Ishitani.
\newblock Estimating the time after death on the basis of corneal opacity.
\newblock {\em Journal of Forensic Research}, 6(1):269, 2015.

\bibitem{canturk2017investigation}
{\.I}smail Cant{\"u}rk, Safa {\c{C}}elik, M~Feyzi {\c{S}}ahin, Fatih Ya{\u{g}}mur, Sad{\i}k Kara, and Fethullah Karabiber.
\newblock Investigation of opacity development in the human eye for estimation of the postmortem interval.
\newblock {\em Biocybernetics and Biomedical Engineering}, 37(3):559--565, 2017.

\bibitem{trokielewicz2020post}
Mateusz Trokielewicz, Adam Czajka, and Piotr Maciejewicz.
\newblock Post-mortem iris recognition with deep-learning-based image segmentation.
\newblock {\em Image and Vision Computing}, 94:103866, 2020.

\bibitem{Czajka2023software}
Adam Czajka, Dennis~J. Chute, Arun Ross, Patrick~J. Flynn, and Kevin~W. Bowyer.
\newblock Software tool and methodology for enhancement of unidentified decedent systems with post-mortem automatic iris recognition.
\newblock {\em New York, 2019-2021. Inter-university Consortium for Political and Social Research [distributor]}, 2023.

\bibitem{Bhuiyan_WACVW_2024}
Rasel~Ahmed Bhuiyan and Adam Czajka.
\newblock Forensic iris image synthesis.
\newblock {\em {Proceedings of the IEEE/CVF Winter Conference on Applications of Computer Vision}}, pages 1015--1023, 2024.

\bibitem{simonyan2014very}
Karen Simonyan and Andrew Zisserman.
\newblock Very deep convolutional networks for large-scale image recognition.
\newblock {\em arXiv preprint arXiv:1409.1556}, 2014.

\bibitem{huang2017densely}
Gao Huang, Zhuang Liu, Laurens Van Der~Maaten, and Kilian~Q Weinberger.
\newblock Densely connected convolutional networks.
\newblock In {\em Proceedings of the IEEE Conference on Computer Vision and Pattern Recognition}, pages 4700--4708, 2017.

\bibitem{he2016deep}
Kaiming He, Xiangyu Zhang, Shaoqing Ren, and Jian Sun.
\newblock Deep residual learning for image recognition.
\newblock In {\em Proceedings of the IEEE Conference on Computer Vision and Pattern Recognition}, pages 770--778, 2016.

\bibitem{szegedy2016rethinking}
Christian Szegedy, Vincent Vanhoucke, Sergey Ioffe, Jon Shlens, and Zbigniew Wojna.
\newblock Rethinking the inception architecture for computer vision.
\newblock In {\em Proceedings of the IEEE Conference on Computer Vision and Pattern Recognition}, pages 2818--2826, 2016.

\bibitem{dosovitskiy2020image}
Alexey Dosovitskiy.
\newblock An image is worth 16x16 words: Transformers for image recognition at scale.
\newblock {\em arXiv preprint arXiv:2010.11929}, 2020.

\bibitem{Trokielewicz_IVC_2020}
Mateusz Trokielewicz, Adam Czajka, and Piotr Maciejewicz.
\newblock Post-mortem iris recognition with deep-learning-based image segmentation.
\newblock {\em Image and Vision Computing}, 94:103866, 2020.

\bibitem{Trokielewicz_WACV_2020}
Mateusz Trokielewicz, Adam Czajka, and Piotr Maciejewicz.
\newblock Post-mortem iris recognition resistant to biological eye decay processes.
\newblock In {\em 2020 IEEE Winter Conference on Applications of Computer Vision (WACV)}, pages 2296--2304, 2020.

\end{thebibliography}
